\documentclass[10pt,journal,compsoc]{IEEEtran}
\usepackage{times}
\usepackage{epsfig}
\usepackage{graphicx}
\usepackage{amsmath}
\usepackage{amssymb}
\usepackage{multirow}
\usepackage{graphicx}
\usepackage{subfigure}
\usepackage{color}

\usepackage{booktabs}
\usepackage{tabulary}
\usepackage{rotating}
\usepackage{bbding}

\usepackage[pagebackref=true,breaklinks=true,letterpaper=true,colorlinks,bookmarks=false]{hyperref}
\newcommand{\bx} {{\bf x }}

\newcommand{\bh} {{\bf h }}

\newcommand{\by} {{\bf y }}

\newcommand{\bI} {{\bf I }}

\newcommand{\etal}{\emph{et al.}}
\newcommand{\eg}{\emph{e.g.}}
\newcommand{\ie}{\emph{i.e.}}

\DeclareMathOperator*{\argmax}{argmax}

\ifCLASSOPTIONcompsoc
  \usepackage[nocompress]{cite}
\else
  \usepackage{cite}
\fi
\ifCLASSINFOpdf
\else
\fi
\begin{document}
%
\title{Faceness-Net: Face Detection through \\ Deep Facial Part Responses}
%
%
%
%

\author{Shuo Yang,
        Ping Luo,
        Chen Change Loy,~\IEEEmembership{Senior Member,~IEEE} 
        and Xiaoou Tang,~\IEEEmembership{Fellow,~IEEE}
\IEEEcompsocitemizethanks{\IEEEcompsocthanksitem Department of Information Engineering, The Chinese University of Hong Kong.\protect\\
E-mail: \{ys014, pluo, ccloy, xtang\}@ie.cuhk,edu.hk
\IEEEcompsocthanksitem Corresponding author: Chen Change Loy
}
}

%
%

\markboth{}%
{Shell \MakeLowercase{\textit{et al.}}: Bare Demo of IEEEtran.cls for Computer Society Journals}
%



\IEEEtitleabstractindextext{%
\begin{abstract}
We propose a deep convolutional neural network (CNN) for face detection leveraging on facial attributes based supervision.
We observe a phenomenon that part detectors emerge within CNN trained to classify attributes from uncropped face images, without any explicit part supervision.
The observation motivates a new method for finding faces through scoring facial parts responses by their spatial structure and arrangement. The scoring mechanism is data-driven, and carefully formulated considering challenging cases where faces are only partially visible. This consideration allows our network to detect faces under severe occlusion and unconstrained pose variations.
Our method achieves promising performance on popular benchmarks including FDDB, PASCAL Faces, AFW, and WIDER FACE. 

\end{abstract}

}

\maketitle

\IEEEdisplaynontitleabstractindextext

%
\IEEEpeerreviewmaketitle

\IEEEraisesectionheading{\section{Introduction}\label{sec:introduction}}

\IEEEPARstart{F}ace detection is an important and long-standing problem in computer vision. A number of methods have been proposed in the past, including neural network based methods~\cite{vaillant1994original,rowley1998neural,garcia2004convolutional,osadchy2007synergistic}, cascade structures~\cite{JointCascade,huang2007high,SURF,viola2004robust} and deformable part models (DPM)~\cite{HeadHunter,yan2014face,zhu2012face} detectors. 
There has been a resurgence of interest in applying convolutional neural networks (CNN) on this classic problem~\cite{qin2016joint,chen2016supervised,jiang2016face,hu2016finding}. Many of these methods follow a cascade object detection framework~\cite{cascadecnn}, some of which directly adopt the effective generic object detection framework RCNN~\cite{RCNN} and Faster-RCNN~\cite{renNIPS15fasterrcnn} as the backbone network, with very deep networks (\eg,~101-layer ResNet) to leverage the remarkable representation learning capacity of deep CNN~\cite{hu2016finding}.  

\begin{figure}[t]
\begin{center}
\includegraphics[width=\linewidth]{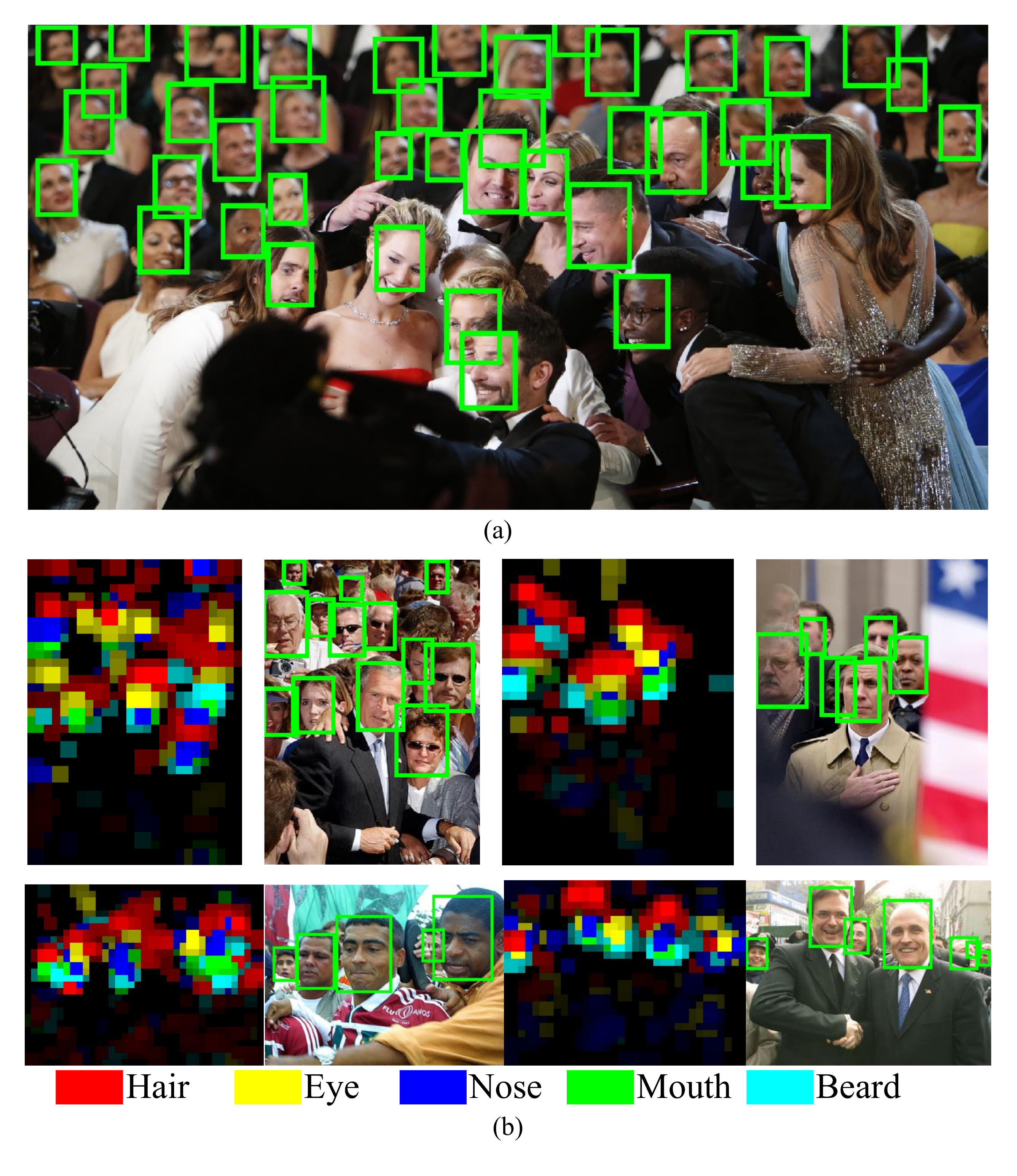}
\vskip -0.6cm
\caption{(a) We propose a deep convolutional network for face detection, which achieves high recall of faces even under severe occlusions and head pose variations. The key to the success of our approach is the new mechanism for scoring face likeliness based on deep network responses on local facial parts. (b) The part-level response maps (we call it `partness' map) generated by our deep network given a full image without prior face detection. All these occluded faces are difficult to handle by conventional approaches.}
\label{fig:intro}
\end{center}
\vskip -0.15cm
\end{figure}

While face bounding boxes have been used as a standard supervisory source for learning a face detector, the usefulness of facial attributes remains little explored.
In this study, we show that facial attributes based supervision can effectively enhance the capability of a face detection network in handling severe occlusions.
As depicted in Fig.~\ref{fig:intro}, a CNN supervised with facial attributes can detect faces even when more than half of the face region is occluded. In addition, the CNN is capable of detecting faces with large pose variation, \eg,~profile view without training separate models under different viewpoints. Such compelling results are hard to achieve by using supervision based on face bounding boxes alone, especially when the training dataset has limited scene diversity and pose variations. 

In this study, we show the benefits of facial attributes supervision through the following considerations:

\noindent
(1) \textit{Discovering facial parts responses supervised by facial attributes}:
The human face has a unique structure. We believe the reasoning of the unique structure of local facial parts (\eg,~eyes, nose, mouth) help detecting faces under unconstrained environments. 
We observe an interesting phenomenon that one can actually obtain part detectors within a CNN by training it to classify part-level binary attributes (\eg,~mouth attributes including big lips, opened mouth, smiling, wearing lipstick) from uncropped face images, without any explicit part supervision.
The trained CNN is then capable of generating high-quality facial part responses in its deep layers that strongly indicate the locations of the face parts.
The examples depicted in Fig.~\ref{fig:intro}(b) show the response maps (known as `partness map' in our paper) of five different face parts.

\noindent
(2) \textit{Computing faceness score from responses configurations}:
Given the parts' responses, we formulate an effective method to reason the degree of face likeliness (which we call \textit{faceness score}) through analyzing their spatial arrangement. For instance, the hair should appear above the eyes, and the mouth should only appear below the nose. Any inconsistency would be penalized. 
Faceness scores will be derived and used to re-rank candidate windows\footnote{There are many options to generate candidate windows. We show two options in this study: (i) using generic object proposal generator, and (ii) using a template proposal. See Sec.~\ref{subsec:region_proposal} for details.} to obtain a set of face proposals.  
Our face proposal approach enjoys a high recall with just a modest number of proposals (over 90\% of face recall with around $150$ proposals, $\approx$0.5\% of full sliding windows, and $\approx$10\% of generic object proposals~\cite{mcg}, measured on the FDDB dataset~\cite{fddb}).

\noindent
(3) \textit{Refining the face hypotheses} --
Both the aforementioned components offer a chance to find a face even under severe occlusion and pose variations.
The output of these components is a small set of high-quality face bounding box proposals that cover most faces in an image.
Given the face proposals, we design a multi-task CNN~\cite{zhang2014facial} in the second stage to refine the hypotheses further, by simultaneously recognizing the true faces and estimating more precise face locations.

Our main contribution in this study is the novel use of CNN and attributes supervision for discovering facial parts' responses. We show that part detectors emerge within a CNN trained to classify attributes from uncropped face images, without any explicit part supervision. The parts' responses are subsequently employed to generate high-quality proposals for training a face detector that is robust to severe occlusion.
The findings aforementioned are new in the literature. 
It is worth pointing out that our network is trained on datasets that are not targeted for face detection (CelebA~\cite{liu2015faceattributes} for face recognition, and AFLW~\cite{aflw} for face alignment) and with simple backgrounds. Nevertheless, it still achieves promising performance on various face detection benchmarks including FDDB, PASCAL Faces, AFW, and the challenging WIDER FACE dataset.

In comparison to our earlier version of this work~\cite{yang2015faceness,loy2017deep}, we present a more effective design of CNN to achieve improved performance and speed.
Firstly, in contrast to our previous work that requires independent convolutional networks for learning responses of different facial parts, we now share feature representations between these attribute-aware networks. The sharing of low and mid-levels representations largely reduce the number of parameters in our framework ($\sim$83\% fewer parameters), while improving the robustness of the feature representation.
Secondly, our previous framework relies on external generic object proposal generators such as selective search~\cite{uijlings2013selective} and EdgeBox~\cite{edgebox} for proposing candidate windows. Inspired by region proposal network presented in~\cite{renNIPS15fasterrcnn}, in this study we directly generate proposals from our attribute-aware networks, thus proposal generation becomes an inherent part of the framework. This design not only leads to improved computation efficiency but also higher recall rate compared with generic object proposal algorithms.
Thirdly, we compare our face detector pre-trained on the task of facial attributes classification with that pre-trained on ImageNet large-scale object classification. 
Apart from the above major changes, we also provide more technical details and discussions. Additional experiments are conducted on the challenging WIDER FACE dataset~\cite{yang2016wider}.

\section{Related Work}
\label{sec:related_work}

There is a long history of using neural network for the task of face detection~\cite{vaillant1994original,rowley1998neural,garcia2004convolutional,osadchy2007synergistic}.
An early face detection survey~\cite{yang2002detecting} provides an extensive coverage on relevant methods. Here we highlight a few notable studies.
Rowley~\etal~\cite{rowley1998neural} exploit a set of neural network-based filters to detect the presence of faces in multiple scales and merge the detections from individual filters.
Osadchy~\etal~\cite{osadchy2007synergistic} demonstrate that a joint learning of face detection and pose estimation significantly improves the performance of face detection.
The seminal work of Vaillant~\etal~\cite{vaillant1994original} adopt a two-stage coarse-to-fine detection. Specifically, the first stage approximately locates the face region, whilst the second stage provides a more precise localization.
Our approach is inspired by these studies, but we introduce innovations on many aspects.
For instance, 
our first stage network is conceptually different from that of~\cite{vaillant1994original}, and many recent deep learning detection frameworks -- we train attribute-aware networks to achieve precise localization of facial parts and exploit their spatial structure for inferring face likeliness. This concept is new and it allows our model to detect faces under severe occlusion and pose variations.
While great efforts have been devoted to addressing face detection under occlusion~\cite{lin2005robust,lin2004fast}, these methods are all confined to frontal faces. In contrast, our model can discover faces under variations of both pose and occlusion.

In the last decades, cascade based~\cite{JointCascade,huang2007high,SURF,viola2004robust} and deformable part models (DPM) detectors dominate face detection approaches. Viola and Jones~\cite{viola2004robust} introduced fast Haar-like features computation via integral image and boosted cascade classifier. Various studies thereafter follow a similar pipeline. Among the variants, SURF cascade~\cite{SURF} was one of the top performers.
Later Chen~\etal~\cite{JointCascade} demonstrate state-of-the-art face detection performance by learning face detection and face alignment jointly in the same cascade framework.
Deformable part models define face as a collection of parts. Latent Support Vector Machine is typically used to find the parts and their relationships. DPM is shown more robust to occlusion than the cascade based methods.
A recent study~\cite{HeadHunter} demonstrates good performance with just a vanilla DPM, achieving better results than more sophisticated DPM variants~\cite{yan2014face,zhu2012face}.

\begin{figure*}[t]
\begin{center}
{\includegraphics[width=0.9\linewidth]{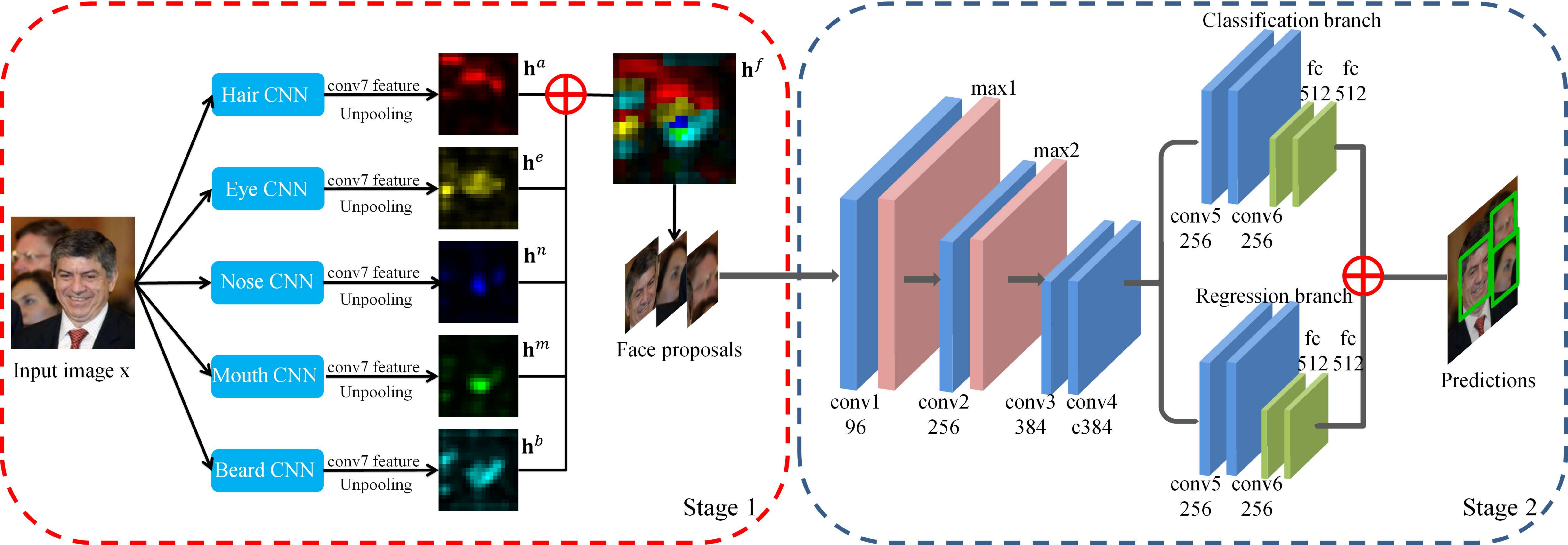}}
\vskip -0.35cm
{\caption{The pipeline of the baseline \textit{Faceness-Net}. The first stage of \textit{Faceness-Net} applies attribute-aware networks to generate response maps of different facial parts. The maps are subsequently employ to produce face proposals. The second stage of \textit{Faceness-Net} refines candidate window generated from first stage using a multi-task convolutional neural network (CNN), where face classification and bounding box regression are jointly optimized. (Best viewed in color).}\label{fig:baseline_facenessnet}}
\end{center}
\vspace{-0.3cm}
\end{figure*}

\begin{figure*}[t]
\begin{center}
{\includegraphics[width=0.9\linewidth]{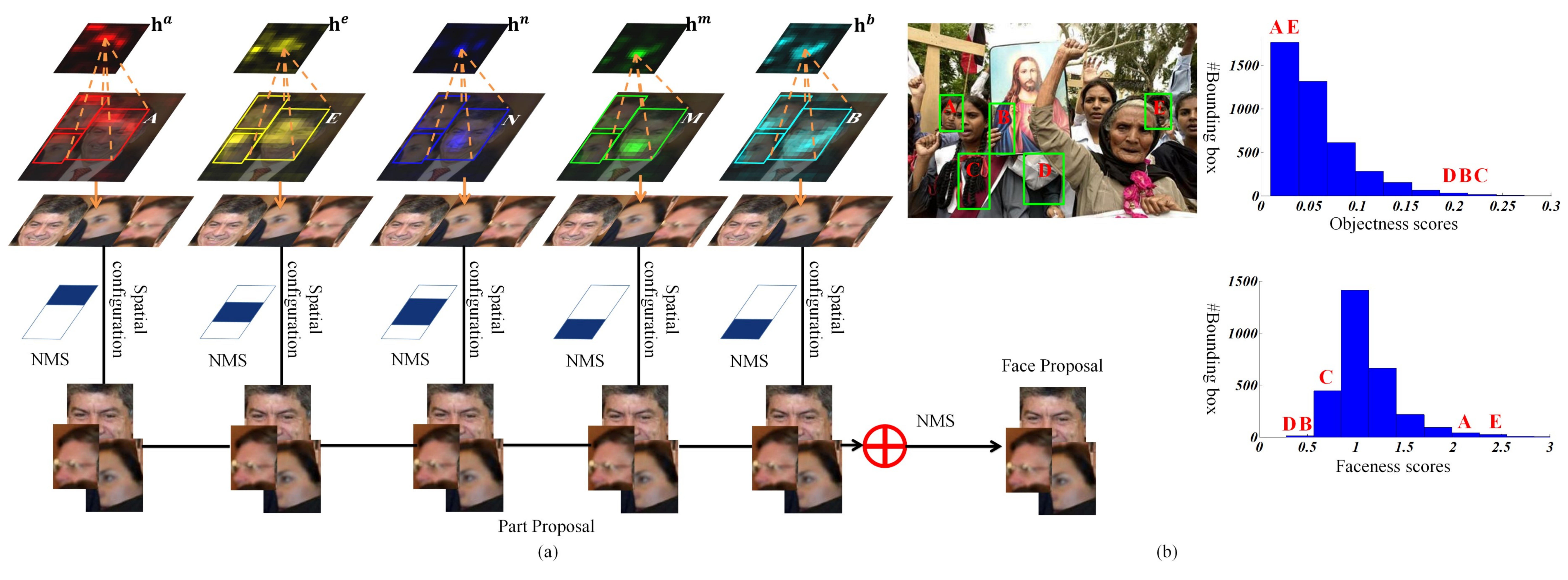}}
\vskip -0.35cm
{\caption{(a) The pipeline for generating face proposals. (b) Bounding box re-ranking by face measure (Best viewed in color).}\label{fig:pipeline_2}}
\vspace{-0.35cm}
\end{center}
\end{figure*}

Recent studies~\cite{farfade2015multi,cascadecnn,chen2016supervised,li2016face3d,opitz16gridloss,yang2017face} show that face detection can be further improved by using deep learning. 
The network proposed by~\cite{farfade2015multi} does not have an explicit mechanism to handle occlusion, the face detector therefore fails to detect faces with heavy occlusions, as acknowledged by the authors. Cascade based convolutional neural networks~\cite{cascadecnn,qin2016joint} replace boosting classifiers with a set of small CNNs to quickly reject negative samples in the early stage. 
Recent studies~\cite{li2016face3d,chen2016supervised} exploit facial landmarks as supervision signals to improve face detection performance. 
In this study, we show that facial attributes can serve as an important source too for learning a robust face detector.

The first stage of our model is partially inspired by generic object proposal approaches~\cite{arbelaez2014multiscale,uijlings2013selective,zitnick2014edge}.
Generic object proposal generators are commonly used in standard object detection algorithms for providing high-quality and category-independent bounding boxes.
These methods typically involve redundant computations over regions that are covered by multiple proposals. 
To reduce computation, Ren~\etal~\cite{renNIPS15fasterrcnn} propose Region Proposal Network (RPN) to generate proposals from high-level response maps in a CNN through a set of predefined anchor boxes. 
Both generic object proposal and RPN methods do not consider the unique structure and parts on the face. Hence, no mechanism is available to recall faces when the face is only partially visible.
These shortcomings motivate us to formulate the new faceness measure to achieve high recall on faces while reducing the number of candidate windows to half the original (compared to the original RPN~\cite{renNIPS15fasterrcnn}).

\section{Faceness-Net}
\label{sec:methodology}

This section introduces the baseline Faceness-Net.
We first briefly overview the entire pipeline and then discuss the details.
As shown in Fig.~\ref{fig:baseline_facenessnet}, Faceness-Net consists of two stages, \ie,~(i) generating face proposals from partness maps by ranking candidate windows using faceness scores, and (ii) refining face proposals for face detection.

\noindent
\textbf{First stage}. A full image $\bx$ is used as an input to a CNN to generate the partness map for each face part.
A set of CNNs, known as attribute-aware networks, are used to generate the partness map of different parts.
The partness map is obtained by weighted averaging over all the response maps at its top convolutional layer. The map indicates the location of a specific facial component presented in the image, \eg,~hair, eyes, nose, mouth, and beard denoted by $\bh^a$, $\bh^e$, $\bh^n$, $\bh^m$, and $\bh^b$, respectively.
For illustration, we sum all these maps into a face label map $\bh^f$, which clearly suggests faces' locations.

Given a set of candidate windows $\{ w \}$ that are generated by existing object proposal methods such as \cite{arbelaez2014multiscale,uijlings2013selective,zitnick2014edge}, or a region proposal network (RPN)~\cite{renNIPS15fasterrcnn},
we rank these windows according to their faceness scores, $\Delta_w$, which are derived from the partness maps with respect to different facial parts configurations, as illustrated at the bottom of Fig.~\ref{fig:pipeline_2}(a).
For example, as visualized in Fig.~\ref{fig:pipeline_2}(a), a candidate window `A' covers a local region of $\bh^a$ (\ie,~hair) and its faceness score is calculated by dividing the values at its upper part with respect to the values at its lower part, because hair is more likely to present at the top of a face region.
The bottom part of Fig.~\ref{fig:pipeline_2}(a) illustrates the spatial configurations of five facial parts. The facial configurations can be learned from the training data.
To reduce the number of the proposed windows, we apply non-maximum suppression (NMS) to smooth the scores by leveraging the spatial relations among these windows.
A final faceness score of `A' is obtained by averaging over the scores of these parts.
We perform another round of NMS to further reduce the number of proposed windows using faceness score.
In this case, a large number of false positive windows can be pruned.
The proposed approach is capable of coping with severe face occlusions. As shown in Fig.~\ref{fig:pipeline_2}(b), face windows `A' and `E' can be retrieved by objectness~\cite{alexe2012measuring} only if a lot of windows are proposed, while windows `A' and `E' rank top $50$ by using our method. 

\noindent
\textbf{Second stage}. The face proposals are refined by training a multi-task CNN, where face classification and bounding box regression are jointly optimized (Fig.~\ref{fig:baseline_facenessnet}).

\subsection{Attribute-Aware Networks}
\label{subsec:attribute_aware_networks}

The first stage of the baseline Faceness-Net consists of multiple attribute-aware networks for generating response maps of different parts (Fig.~\ref{fig:baseline_facenessnet}). Five networks are needed to cover all five pre-defined facial components, \ie,~hair, eyes, nose, mouth, and beard. These attribute-aware networks share the same structure. Next, we first discuss the network structure and subsequently show that these networks can share representation to reduce parameters.

\begin{figure}[t]
\begin{center}
\vskip -0.3cm
\includegraphics[width=\linewidth]{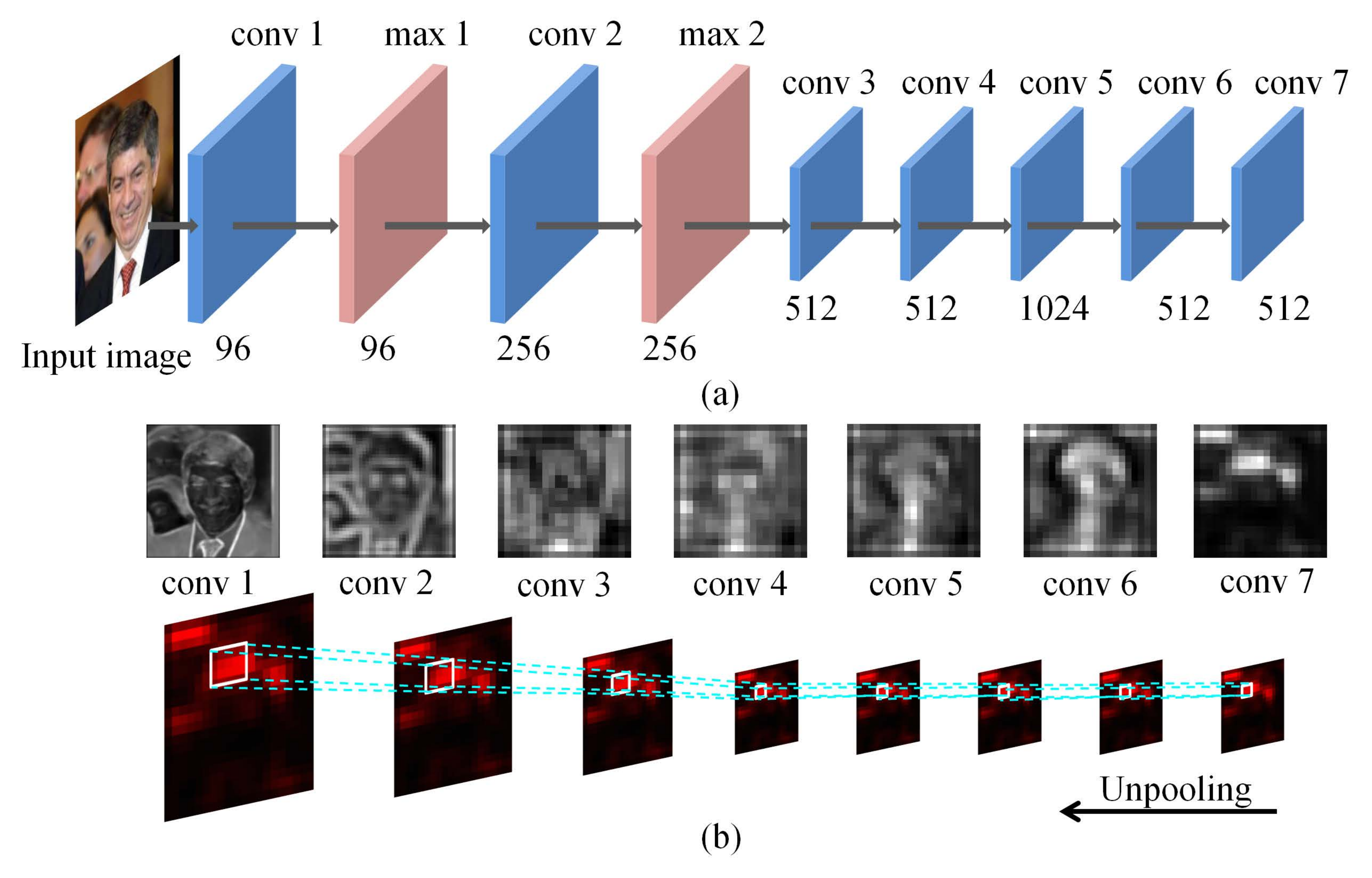}
\caption{In the baseline Faceness-Net, we adopt different attribute-aware networks for different facial parts. (a) This figure shows the architecture of an attribute-aware deep network used for discovering the responses of `hair' component. Other architectures are possible. See Sec.~\ref{subsec:attribute_aware_networks} for details. (b) The response map from conv7, which is generated by applying element-wise averaging along the channels for $l2$ normalized feature maps, indicates the location of hair component. The response map is upsampled through unpooling operation~\cite{clarifai14} to obtain the final partness map of the same size as the input image.}
\label{fig:attribute_aware_network}. 
\end{center}
\end{figure}

\noindent\textbf{Network structure}. 
The choice of network structure for extracting partness maps is flexible. Figure~\ref{fig:attribute_aware_network}(a) depicts the structure and hyper-parameters of the CNN used in the baseline Faceness-Net. 
This convolutional structure is inspired by the AlexNet~\cite{imagenet}, which was originally proposed for object categorization. 
Specifically, the network stacks seven convolutional layers (conv1 to conv7) and two max-pooling layers (max1 and max2). The hyper-parameters of each layer is specified in Fig.~\ref{fig:attribute_aware_network}(a). 

Once the attribute networks are trained (training details are provided in Sec.~\ref{subsec:learning_partness}), we obtain the response map of a part through first applying $l2$ normalization on the feature map for each channel and then element-wise averaging along the channels.
We examine the response maps obtained from the attribute-aware networks. As observed from Fig.~\ref{fig:attribute_aware_network}(b), the feature maps of the first few convolutional layers do not clearly indicate the locations of facial parts. However, a clear indication of the facial component can be seen from responses of conv7.
Consequently, we obtain an initial response map from the conv7 layer. The final partness map that matches the input image's size is obtained through performing unpooling~\cite{clarifai14} on the conv7's response map.

\begin{figure}[t]
\begin{center}
\includegraphics[width=\linewidth]{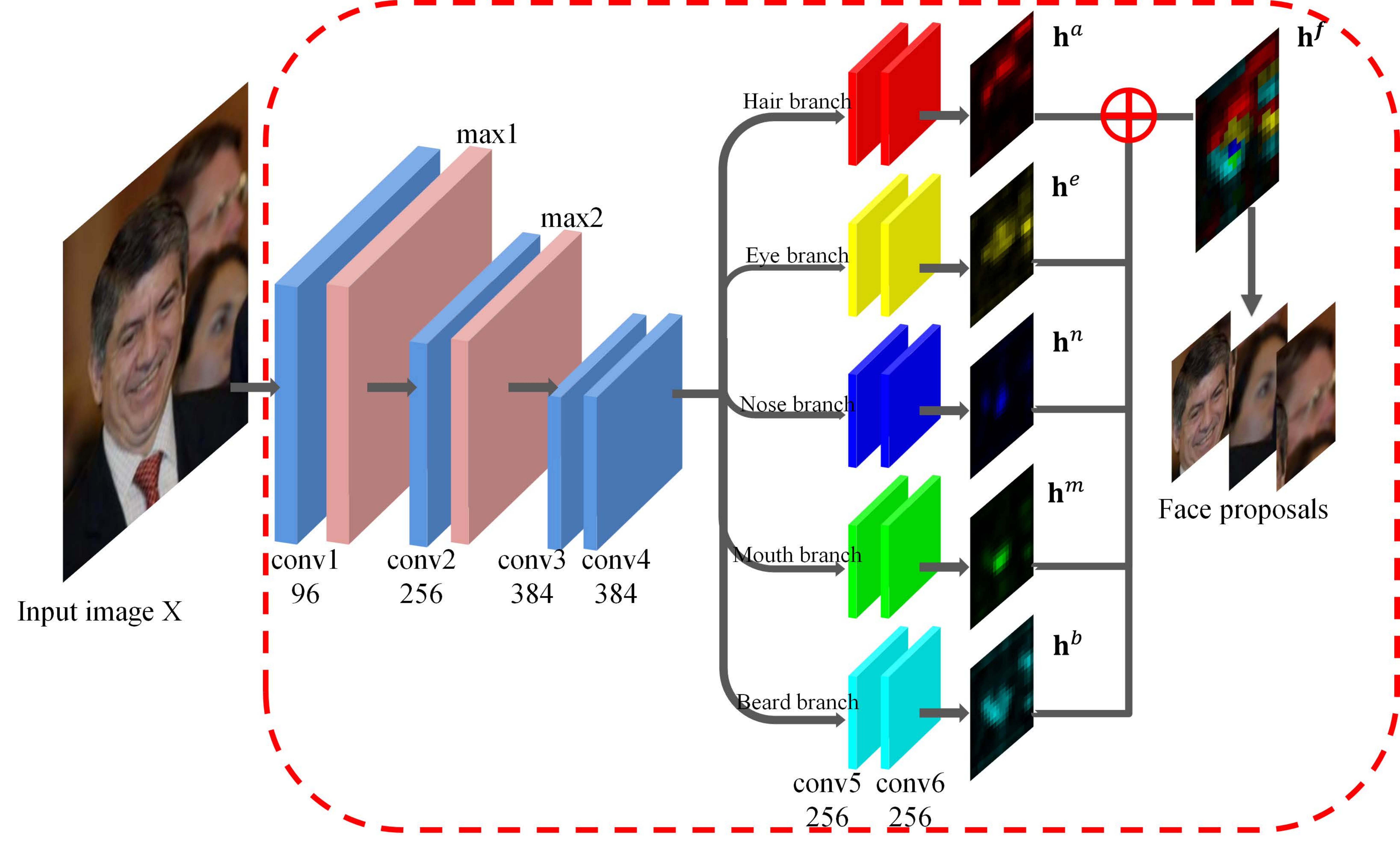}
\vskip -0.3cm
\caption{The first stage of Faceness-Net-SR, a variant of the baseline Faceness-Net. We share representations between the different attribute-aware networks and reduce filters leading to improved efficiency and performance. See Sec.~\ref{subsec:attribute_aware_networks} for details.}
\label{fig:shared_representation}
\end{center}
\end{figure}

\noindent\textbf{Shared representation}.
It is observed that the feature maps of earlier layers across the different attribute-aware networks are almost identical and they are not indicative of parts' locations.
Motivated by these observations, instead of designating separate attribute-aware networks for different facial components, we share early convolutional layers of these networks to reduce parameters. Specifically, the first four convolutional layers that do not clearly suggests parts' locations are shared, followed by five branches, each of which consists of two convolutional layers responsible for a facial component, as shown in Fig.~\ref{fig:shared_representation}.
Note that in comparison to the structure presented in Fig.~\ref{fig:attribute_aware_network}(a), we additionally remove a convolutional layer and trim the number of filters in other layers to reduce parameters.
The sharing of representation and filter reduction lead to a single attribute-aware network with $83\%$ fewer parameters than the original five attribute-aware networks.
We denote a Faceness-Net with shared representation as Faceness-Net-SR.
We will show that this network structure not only reduces computations but also improves the robustness of feature representation for face detection.

\subsection{Learning to Generate Partness Maps}
\label{subsec:learning_partness}

\noindent\textbf{Pre-training the attribute-aware networks}.
Pre-training generally helps to improve the performance of a deep network.
There are two plausible pre-training options depending upon whether we share the representations across attribute-aware networks or not.

The first option is to pre-train our attribute-aware networks with massive general object categories in ImageNet~\cite{russakovsky2015imagenet}. 
From our observations, this option works well when the representations across networks are not shared. Since each attribute-aware network originally has access only to a particular group of data specific to a certain attribute, the larger-scale ImageNet data helps to mitigate the overfitting issue that is caused by insufficient data.

The second option omits the ImageNet pre-training stage and trains a network directly on the task of facial attributes classification. This option works best when we adopt the shared representation scheme discussed in Sec.~\ref{subsec:attribute_aware_networks}. Thanks to the sharing of representation, the attribute-aware network requires a relatively smaller quantity of training data. Thus, no overfitting is observed despite we use the facial attributes dataset, which is much smaller in scale, \ie,~180,000 images compared to 1 million images in ImageNet.

\noindent\textbf{Fine-tuning the attribute-aware networks}.
Once an attribute-network is pre-trained, we can fine-tune it to generate the desired partness maps. There are different fine-tuning strategies, but not all of them can generate meaningful partness maps for deriving a robust faceness score.

\begin{figure}[t]
\begin{center}
\includegraphics[width=\linewidth]{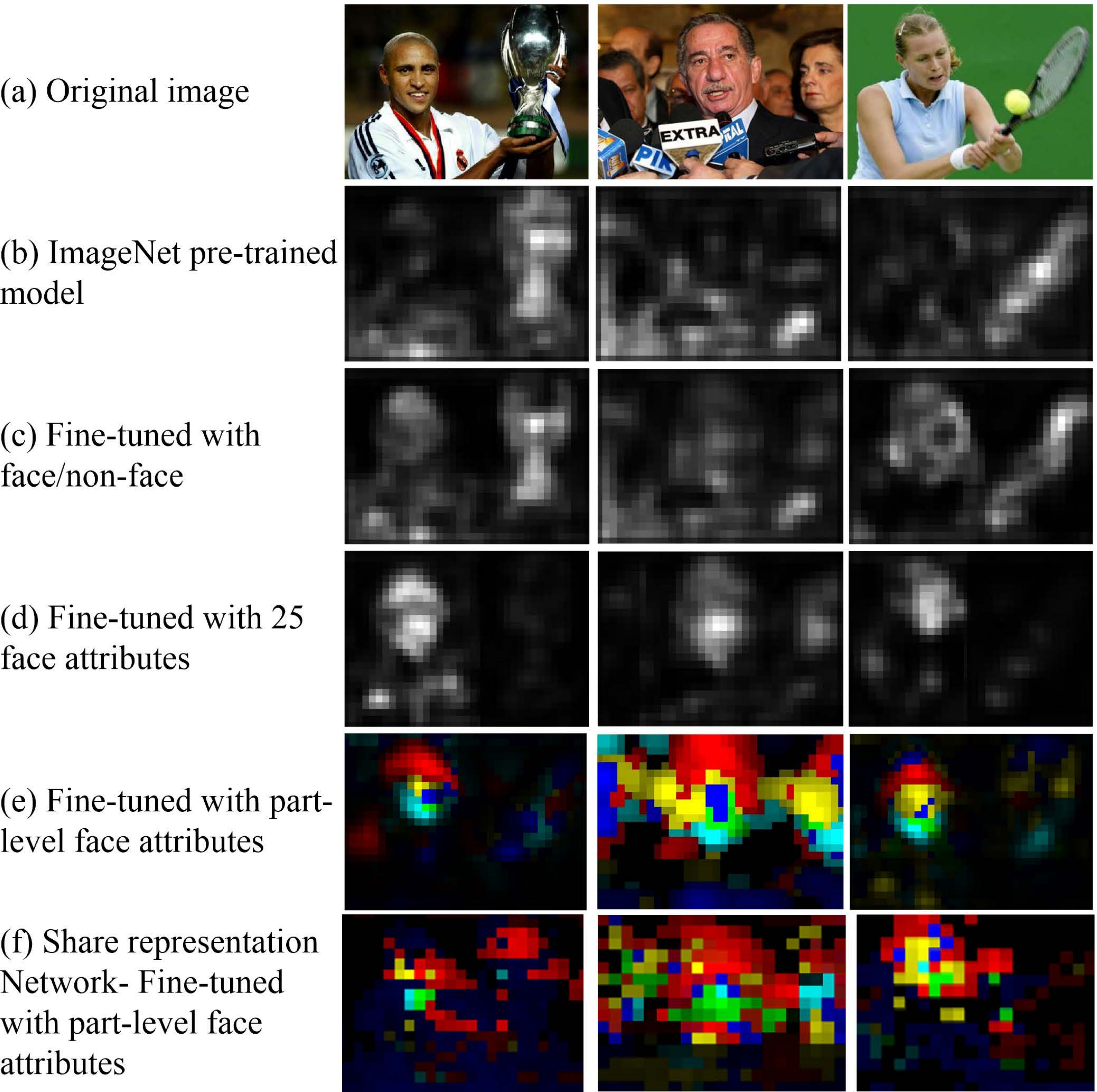}
\vskip -0.2cm
\caption{The partness maps obtained by using different types of supervisions and fine-tuning strategies. The maps in (a-e) are generated using the baseline Faceness-Net depicted in Fig.~\ref{fig:baseline_facenessnet}. The maps in (f) is generated using Faceness-Net-SR with shared representation, as illustrated in Fig.~\ref{fig:shared_representation}.}
\label{fig:partness_map_learning}
\end{center}
\end{figure}

As shown in Fig.~\ref{fig:partness_map_learning}(b), a deep network trained on generic objects, \eg,~AlexNet~\cite{imagenet}, is not capable of providing us with precise faces' locations, let alone partness map.
To generate accurate partness maps, we explore multiple ways for learning an attribute-aware network.
The most straightforward manner is to use the image and its pixel-wise segmentation label map as input and target, respectively.
This setting is widely employed in image labeling~\cite{farabet2013learning,aerial_hinton}.
However, it requires label maps with pixel-wise annotations, which are expensive to collect.
Another setting is image-level classification (\ie,~faces and non-faces), as shown in Fig.~\ref{fig:partness_map_learning}(c). It works well where the training images are well-aligned, such as face recognition~\cite{sun2014deepb}.
Nevertheless, it suffers from complex background clutter because the supervisory information is not sufficient to account for rich and diverse face variations.
Its learned feature maps contain too many noises, which overwhelm the actual faces' locations.
Attribute learning in Fig.~\ref{fig:partness_map_learning}(d) extends the binary classification in (c) to the extreme by using a combination of attributes to capture face variations.
For instance, an `Asian' face can be distinguished from a `European' face.
However, our experiments demonstrate that this setting is not robust to occlusion.

\begin{table}[t]
\begin{center}
\caption{Facial attributes grouping.}
\label{tab:attribute_category}
\vskip -0.25cm
\scriptsize
\addtolength{\tabcolsep}{-1pt}
\begin{tabular}{c||c}
\hline
\textbf{Facial Part} & \textbf{Facial Attributes}
\\
\hline
\hline
Hair & Black hair, Blond hair, Brown hair, Gray hair, Bald, \\& Wavy hair, Straight hair, Receding hairline, Bangs\\
\hline
Eye & Bushy eyebrows, Arched eyebrows, Narrow eyes,\\ & Bags under eyes, Eyeglasses \\
\hline
Nose & Big nose, Pointy nose \\
\hline
Mouth &	Big lips, Mouth slightly open, Smiling,\\ & Wearing lipstick\\
\hline
Beard &	No beard, Goatee, 5 o'clock shadow, \\ & Mustache, Sideburns\\
\hline
\end{tabular}
\end{center}
\end{table}

Figure~\ref{fig:partness_map_learning}(e) shows the partness maps obtained by the baseline Faceness-Net, for which the attribute networks do not share representations.
The strategy we propose extends (d) by partitioning attributes into groups based on facial components.
For instance, `black hair', `blond hair', `bald', and `bangs' are grouped together, as all of them are related to hair.
The grouped attributes are summarized in Table~\ref{tab:attribute_category}.
In this case, face parts are modeled separately. If one part is occluded, the face region can still be localized by the other parts.
We take the Hair-Branch shown in the stage one of Fig.~\ref{fig:baseline_facenessnet} as an example to illustrate the learning procedure.
Let $\{\bx_i,\by_i\}_{i=1}^N$ be a set of full face images and the attribute labels of hair. Images are first resized to $128\times128$ where $x_i \in \mathbb{R}^{128\times128}$ and $\by_i\in\{0, 1\}^{1\times M}$ indicate there are nine attributes ($M=9$) related to hair as listed in Table~\ref{tab:attribute_category}\footnote{Other target designs~\cite{yang15target} are applicable.}.
Learning is formulated as a multi-variate classification problem by minimizing the cross-entropy loss,
\begin{equation}\label{eq:finetune}
\begin{split}
L=&\sum_{i=1}^{N} \sum_{j=1}^{M}\by_i^j\log p(\mathbf{y}_i^j=1|\bx_i) + \\&
({1} - \by_i^j)\log\big({1} - p(\by_i^j=1|\bx_i)\big),
\end{split}
\end{equation}
where $p(\by_i^j|\bx_i)$ is modeled as a sigmoid function,
\ie~$p(\by_i^j=k|\bx_i)=\frac{1}{1 + \exp(-f(\bx_i))}$,
indicating the probability of the presence of $jth$ attributes.
The features of $\bx_i$ are denoted as $f(\bx_i)$.
To facilitate the learning, we stack two fully-connected layers on top of the last convolutional layer of the structure shown in Fig.~\ref{fig:attribute_aware_network}.
We optimize the loss function by using stochastic gradient descent with back-propagation.
After training the attribute-aware network, the fully-connected layers are removed to make the network fully convolutional again.

Figure~\ref{fig:partness_map_learning}(f) shows the partness maps that are generated from the networks with shared representation, \ie,~Faceness-Net-SR (see Fig.~\ref{fig:shared_representation}). 
Visually, the partness maps generated by this model are noisier compared to Fig.~\ref{fig:partness_map_learning}(e). The key reason is that the Faceness-Net-SR is not pre-trained using ImageNet data but directly trained on the attribute classification task.
Despite the noisy partness maps, they actually capture more subtle parts' responses and therefore lead to higher recall rate in the subsequent face proposal stage, provided that the number of proposals is sufficiently large.

\subsection{Generating Candidate Windows}
\label{subsec:region_proposal}

Face detection can be improved if the inputs are formed by a moderate number of proposals with a high recall rate. 
To produce the required proposals, we will explore two plausible choices to generate the initial set of candidate windows.

\noindent \textbf{Generic object proposal}. Generic object scoring is primarily employed to reduce the computational cost of a detector.
It has also been shown improving detection accuracy due to the reduction of spurious false positives~\cite{alexe2012measuring}.
A variety of cues has been proposed to quantify the objectness of an image window, \eg,~norm of the gradient~\cite{bing}, edges~\cite{zitnick2014edge}, or integration of a number of low-level features~\cite{alexe2012measuring}.
Other popular methods include super-pixel based approaches, \eg,~selective search~\cite{uijlings2013selective}, randomized Prim~\cite{manen2013prime}, and multi-scale combinatorial grouping \cite{arbelaez2014multiscale}.
Our framework can readily employ these generic candidate windows for ranking using the proposed faceness score (Sec.~\ref{subsec:faceness_measure}). 

\begin{figure}[t]
\begin{center}
{\includegraphics[width=\linewidth]{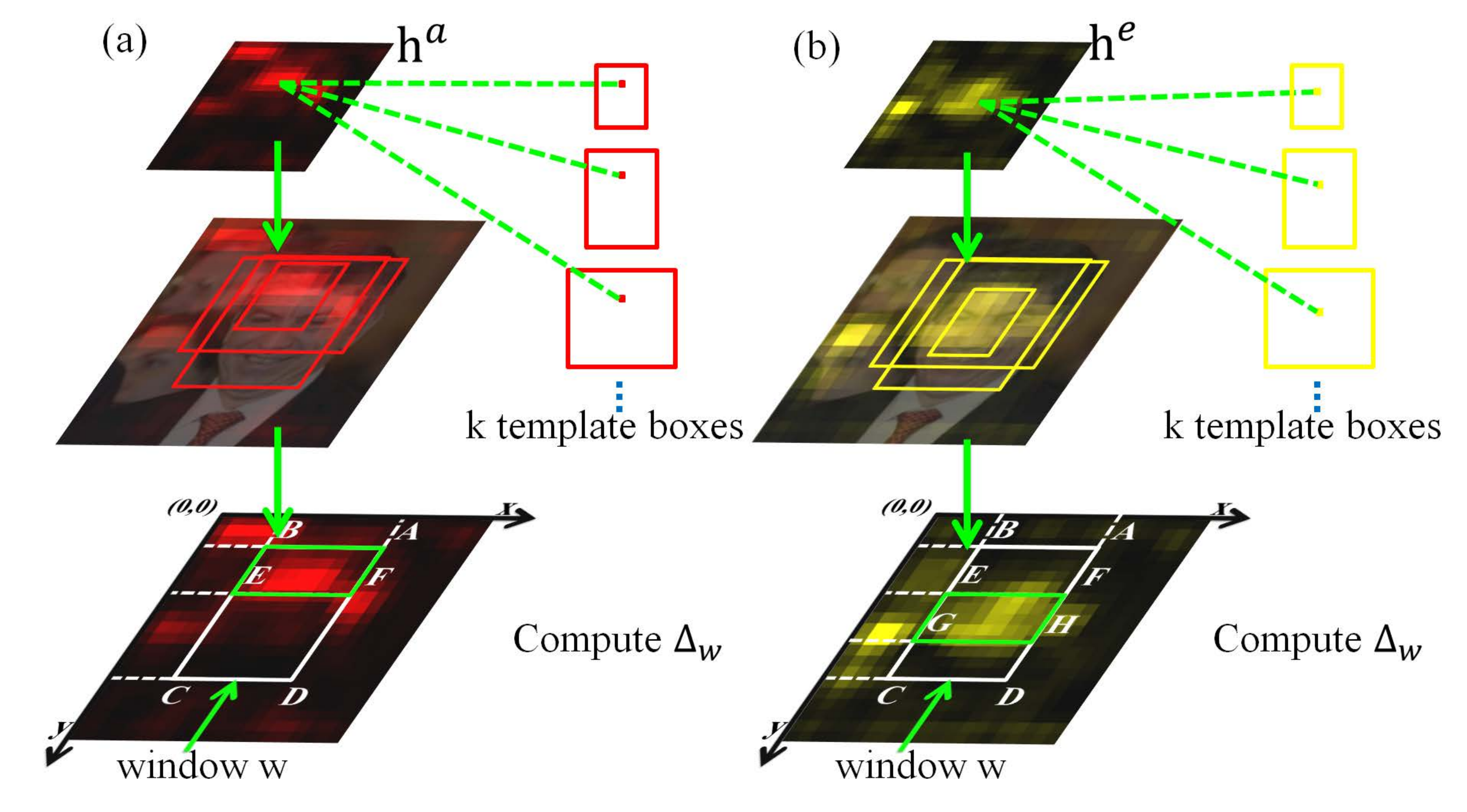}}
\vskip -0.3cm
{\caption{Examples of template proposal and faceness measurement. The partness maps of hair and eyes are shown in (a) and (b), respectively. $\Delta_w$ is the faceness score of a window, $w$. (Best viewed in color).}\label{fig:region_proposal}}
\end{center}
\vskip -0.3cm
\end{figure}

\noindent \textbf{Template proposal}.
In order to decouple the dependence of object proposal algorithms to generate candidate windows, we propose a template proposal method in which candidate windows are generated from multiple predefined templates on feature maps.

We provide an example below on using a partness map of hair, $\bh^a$, for template proposal. As shown in Fig.~\ref{fig:region_proposal}, each value of location $(i,j)$ on the partness map $\bh^a$ indicates the probability of the appearance of the hair component.
We select a set of $M$ locations $\{(h_i, h_j)\}_{i=1}^M$ with a probability $p_{(h_i,h_j)}$ higher than $t$. For each selected location, multiple template proposals are generated, where the number of maximum possible proposals for each location is fixed as $k$.
The proposals are obtained from predefined reference boxes, which we call templates.
For each face part, templates are centered at different locations considering the structure of the human face. In addition, they are associated with a specific scale and aspect ratio, as shown at the top of Fig.~\ref{fig:region_proposal}.
For instance, the templates of the hair region are centered at $(W/2, H/3)$ and the templates of eyes are centered at $(W/2, H/2)$, where $W$ and $H$ represent the width and height of an anchor. 
Similar to previous work~\cite{renNIPS15fasterrcnn}, these templates are translation invariant up to the network's total stride, and the method does not incur extra cost for addressing scales thanks to the multi-scale templates.

In our study, we define $10$ scales and $1$ aspect ratio, yielding $k=10$ templates at each selected position. Specifically, we use $10$ scales with box areas of $25^2$, $50^2$, $75^2$, $100^2$, $135^2$, $170^2$, $200^2$, $240^2$, $300^2$, and $350^2$ pixels, and $1$ aspect ratio of $1:1.5$ (with width to height).
The parameters of templates, \ie,~center location, scale and aspect ratio are selected by maximizing the recall rate given an average number of $n$ proposals per image.
In our study, we perform a grid search on the training set to select the parameters.  

\noindent \textbf{Discussion}.
Both the generic objectness measures and RPN (trained on ImageNet) are devoted to generic objects therefore not suitable to propose windows specific to faces. In particular, applying a generic proposal generator directly would produce an enormous number of candidate windows but an only minority of them contain faces.
While RPN is computationally more efficient than generic object proposal generators, it cannot be directly applied to our problem too.
Specifically, in order for the RPN to cope with faces with tiny size and various poses, a large number of anchor boxes are required, leading to an enormous number of proposals.
In the next section, we discuss a new faceness measure that can complement existing object proposal generators or the template proposal method to achieve high recall on faces, while significantly reduce the number of candidate windows. 
The proposed faceness measure scheme is in practice related to traditional face detector schemes based on Haar features, with the difference that here the Haar features pool CNN feature responses instead of pixel luminance values.

\subsection{Ranking Windows by Faceness Score}
\label{subsec:faceness_measure}

After generating candidate windows based on the methods described in Sec.~\ref{subsec:region_proposal}, our approach computes a faceness score on these windows to return a ranked set of top-scoring face proposals.
Figure~\ref{fig:region_proposal} illustrates the procedure of deriving the faceness measure from the partness maps of hair and eyes.
Let $\Delta_w$ be the faceness score of a window $w$.
For example, as shown in Fig.~\ref{fig:region_proposal}(a), given a partness map of hair, $\bh^a$, $\Delta_w$ is attained by dividing the sum of values in ABEF (green) by the sum of values in FECD.
Similarly, Fig.~\ref{fig:region_proposal}(b) shows that $\Delta_w$ is obtained by dividing the sum of values in EFGH (green) with respect to ABEF+HGCD of $\bh^e$.
For both of the above examples, a larger value of $\Delta_w$ indicates a higher overlapping ratio of $w$ with a face.
The choice of method for computing the faceness score is flexible. It is possible to compute the faceness score using other forms of handcrafted features that can effectively capture the face structure through response maps.

The spatial configurations, such as ABEF in Fig.~\ref{fig:region_proposal}(a) and EFGH in Fig.~\ref{fig:region_proposal}(b), can be learned from data.
We take hair as an example.
We need to learn the positions of points E and F, which can be represented by the $(x,y)$-coordinates of ABCD, \ie,~the proposed window.
For instance, the position of E in Fig.~\ref{fig:region_proposal}(a) can be represented by $x_e=x_b$ and $y_e=\lambda y_b+(1-\lambda)y_c$, implying that the value of its $y$-axis is a linear combination of $y_b$ and $y_c$.
With this representation, $\Delta_w$ can be efficiently computed by using the integral image (denoted as $\bI$) of the partness map. For instance, $\Delta_w$ in (a) is attained by $\frac{\bI_{ABEF}}{\bI_{CDEF}}$.
\begin{equation}
\begin{split}
\bI_{ABEF} &=\bI(x_f, \lambda y_a+(1-\lambda)y_d\big)+\bI(x_b,y_b)\\&-\bI(x_a,y_a)-\bI\big(x_b,\lambda y_b+(1-\lambda)y_c\big)\\
\bI_{CDEF} &=\bI(x_d,y_d)+\bI(x_e,y_e)\\&-\bI\big(x_a,\lambda y_a+(1-\lambda)y_d\big)-\bI(x_c,y_c)
\end{split}
\end{equation}
where $\bI(x,y)$ signifies the value at the location $(x,y)$.

Given a training set $\{w_i,r_i,\bh_i\}_{i=1}^M$, where $w_i$ and $r_i\in\{0,1\}$ denote the $i$-th window and its label (\ie~face/non-face), respectively.
Let $\bh_i$ be the cropped partness map with respect to the $i$-th window, \eg,~region ABCD in $\bh^a$. This problem can be formulated as maximum a posteriori (MAP) estimation
\begin{equation}\label{eq:map}
\lambda^\ast=\argmax_{\lambda}\prod_i^M p(r_i|\lambda,w_i,\bh_i)p(\lambda,w_i,\bh_i),
\end{equation}
where $\lambda$ represents a set of parameters when learning the spatial configuration of hair (Fig.~\ref{fig:region_proposal}(a)).
The terms $p(r_i|\lambda,w_i,\bh_i)$ and $p(\lambda,w_i,\bh_i)$ denote the likelihood and prior, respectively. The likelihood of faceness can be modeled by a sigmoid function, \ie,~$p(r_i|\lambda,w_i,\bh_i)=\frac{1}{1+\exp(\frac{-\alpha}{\Delta_{w_i}})}$, where $\alpha$ is a coefficient.
This likelihood measures the confidence of partitioning the face and non-face, given a certain spatial configuration.
The prior term can be factorized, $p(\lambda,w_i,\bh_i)=p(\lambda)p(w_i)p(\bh_i)$,
where $p(\lambda)$ is a uniform distribution between zero and one, as it indicates the coefficients of linear combination, $p(w_i)$ models the prior of the candidate window, which can be generated by object proposal methods, and $p(\bh_i)$ is the partness map as obtained in Sec.~\ref{subsec:learning_partness}. 
Since $\lambda$ typically has a low dimension (\eg,~one dimension of hair), it can be simply obtained by line search. Note that Eq.~\eqref{eq:map} can be easily extended to model more complex spatial configurations. This process is similar with learning Haar templates using boosting classifier, but requires less computation while achieving good performance compared with more elaborated process.

\subsection{Face Detection}
\label{sec:face_detection}

The top candidate windows that are ranked by faceness score attain a high recall rate. These face proposals can be subsequently fed to the multi-task CNN at stage 2 of the proposed pipeline (Fig.~\ref{fig:baseline_facenessnet}) for face detection.

\noindent \textbf{Pre-training}. We directly use the earlier layers of attribute-aware networks (the stage-1 network with shared representation as shown in Fig.~\ref{fig:shared_representation}) up to conv4 as the pre-trained model for the multi-task CNN of stage 2. After conv4, as shown in Fig.~\ref{fig:baseline_facenessnet}, the multi-task CNN forks into two branches, each of which consists of two convolutional layers and two fully connected layers. The two branches are optimized to handle different tasks, namely face classification and bounding box regression, respectively.

It is worth pointing out that the multi-task CNN can be pre-trained on the ImageNet data, instead of reusing the parameters of the attribute-aware networks. Nevertheless, we found that the multi-task CNN converges much faster given the face attributes based pretrained model. Specifically, the attribute pretrained network only requires $45,000$ iterations to converge during the face detection fine-tuning stage, in comparison to more than $200,000$ iterations for the ImageNet pertrained network using the same mini-batch size. We conjecture that much less effort is needed to transform the feature representations learned from the facial attribute classification task to the face detection task.

\noindent \textbf{Multi-task fine-tuning}.
We fine-tune the first branch of the multi-task CNN for face classification and the second branch for bounding box regresssion.
Fine-tuning is performed using the face proposals obtained from the previous step (Sec~\ref{subsec:faceness_measure}).
For face classification, we assign a face proposal to its closest ground truth bounding box based on the Euclidean distance between their respective center coordinates. 
A face proposal is considered positive if the Intersection over Union (IoU) between the proposal box and the assigned ground truth box is larger than $0.5$; otherwise it is negative.
For bounding box regression, we train the second branch of the multi-task CNN to regress each proposal to the coordinates of its assigned ground truth box.
If the proposed window is a positive sample, the regression target is generated by Eq.~\eqref{eq:regression}. We use the following parameterizations of the 4 coordinates:
\begin{equation}\label{eq:regression}
\begin{split}
& x^{*}_{1}=(x_1-x'_{1}) / \zeta, \;\;\; y^{*}_{1}=(y_1-y'_{1}) / \zeta \\
& x^{*}_{2}=(x_2-x'_{2}) / \zeta, \;\;\; y^{*}_{2}=(y_2-y'_{2}) / \zeta, \\
\end{split}
\end{equation}
where $\zeta = \max(x'_{2}-x'_{1},y'_{2}-y'_{1})$ is a normalizing factor. The vector $[x_1,y_1,x_2,y_2]$ denotes the top-left and bottom-right coordinates of a bounding box. Variables $x$, $x'$,  and $x^*$ represent the ground truth box, proposed box, and regression target. This process normalizes regression target into a range of $[-1,1]$ which can be easily optimized by using least square loss. The standard bounding box regression targets~\cite{renNIPS15fasterrcnn} and $L1$ loss are also applicable.
If a proposed window is non-face, the CNN outputs a vector of $[-1,-1,-1,-1]$ whose gradients will be ignored during back propagation.

More implementation details are given below.
During the training process, if the number of positive samples in a mini-batch is smaller than $20\%$ of the total samples, we randomly crop the ground truth faces and add these samples as additional positive samples. Therefore, the ratio of positive samples and negative samples is kept not lower than $1:4$.
Meanwhile, we conduct bounding box NMS on the negative samples. The IoU for the NMS is set to $0.7$. The proposed bounding boxes are cropped and then resized to $128\times128$. 
To handle blurry faces, we augment our training samples by applying Gaussian blurring. The fine-tuning consumes $50K$ iterations with a batch size of $256$ images.  We adopt Euclidean loss and cross-entropy loss for bounding box regression and face classification, respectively.

\section{Experimental Settings}
\label{sec:settings}

\noindent \textbf{Training datasets}.
\noindent (i) We employ CelebA dataset~\cite{liu2015faceattributes} to train our attribute-aware networks. The dataset contains $202,599$ web-based images exclusive from the LFW~\cite{lfw}, FDDB~\cite{fddb}, AFW~\cite{zhu2012face} and PASCAL~\cite{yan2014face} datasets. Every image in the dataset are labeled with $40$ facial attributes.
We select $25$ facial attributes from CelebA dataset for each image and divide the attributes into five categories based on their respective facial parts as shown in Table~\ref{tab:attribute_category}.
We randomly select $180,000$ images from the CelebA dataset for training and the remaining is reserved as the validation set.
\noindent (ii)
For face detection training, we choose $13,205$ face images from the AFLW dataset~\cite{aflw} to ensure a balanced out-of-plane pose distribution.
We observe a large number of missed annotated faces in the AFLW dataset, which could hamper the training of our face detector. Hence, we re-annotate face bounding boxes for those missing faces.
The total number of faces in the re-annotated AFLW is $29,133$ compared with $24,386$ in the original data.  
As negative samples, we randomly select $5,771$ person-free images from the PASCAL VOC 2007 dataset~\cite{pascalvoc}.

\begin{figure}[t]
\begin{center}
\vskip -0.3cm
{\includegraphics[width=\linewidth]{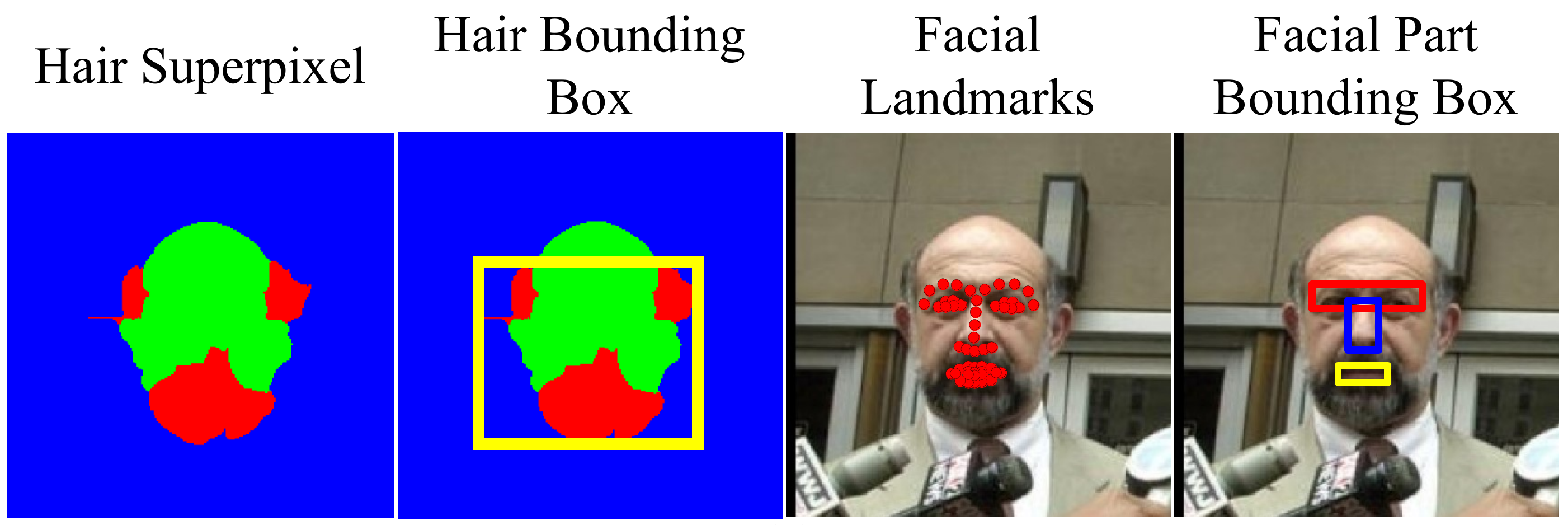}}
{\caption{The figure shows examples of ground truth bounding boxes of facial parts. Hair ground truth bounding boxes are generated from superpixel maps~\cite{crf}. Eye, nose, and mouth bounding boxes are generated from $68$ ground truth facial landmarks~\cite{lfw_dense_landmarks}.}\label{fig:facial_part_gt}}
\vskip -0.7cm
\end{center}
\end{figure}

\noindent \textbf{Part response test dataset}.
We use LFW dataset~\cite{lfw} for evaluating the quality of part response maps for part localization.
Since the original dataset does not come with part-level bounding boxes, we label the boxes with the following scheme. We follow the annotations provided by~\cite{crf} on hairs and beard for a set of $2927$ LFW images. Hair bounding boxes are generated with minimal and maximal coordinates of hair superpixel as shown in Fig.~\ref{fig:facial_part_gt}.
Using a similar strategy, eye, nose and mouth bounding boxes are obtained from the manually labeled $68$ dense facial landmarks~\cite{lfw_dense_landmarks} on the original LFW~\cite{lfw} images, as shown in Fig.~\ref{fig:facial_part_gt}.

\noindent \textbf{Face proposal and detection test datasets}.
We use the following datasets.
\noindent (i) FDDB~\cite{fddb} dataset contains $5,171$ faces in a set of $2,845$ images. For the face proposal evaluation, we follow the standard evaluation protocol widely used in object proposal studies~\cite{zitnick2014edge} and transform the original FDDB ellipses ground truth into bounding boxes by minimal bounding rectangle.
For the face detection evaluation, the original FDDB ellipse ground truth is used.
\noindent (ii) AFW~\cite{zhu2012face} dataset contains $205$ Flickr images with $473$ annotated faces of large variations in both face viewpoint and appearance.
\noindent (iii) PASCAL faces~\cite{yan2014face} is a widely used face detection benchmark dataset. It consists of $851$ images and $1,341$ annotated faces.
\noindent (iv) WIDER FACE~\cite{yang2016wider} is the largest and extremely challenging face detection benchmark dataset. It consists of $32,203$ images and $393,703$ annotated faces.


\noindent \textbf{Evaluation settings}.
Following~\cite{zitnick2014edge}, we employ the Intersection over Union (IoU) as our evaluation metric.
We fix the IoU threshold to $0.5$ following the strict PASCAL criterion. In particular, an object is considered being covered/detected by a proposal if the IoU is no less than $0.5$.
To evaluate the effectiveness of different object proposal algorithms, we use the detection rate (DR) given the number of proposals per image~\cite{zitnick2014edge}.
For face detection, we use standard precision and recall (PR) to evaluate the effectiveness of face detection algorithms.

\noindent \textbf{Faceness-Net Variants}.
We evaluate four variants of Faceness-Net: \vspace{-0.2cm}
\begin{itemize}
\item Faceness-Net - our baseline method mentioned in Sec.~\ref{sec:methodology} with five attribute-aware networks Fig.~\ref{fig:baseline_facenessnet}. An external generic object proposal generator is adopted.
\item Faceness-Net-SR - a variant with a single attribute-aware network by sharing representations, as described in Sec.~\ref{subsec:attribute_aware_networks}. Fig.~\ref{fig:shared_representation} shows the network structure of Faceness-Net-SR. An external generic object proposal generator is adopted.
\item Faceness-Net-TP - a variant of the Faceness-Net that adopts the template proposal technique to generate candidate windows. Details can be found in Sec.~\ref{subsec:region_proposal}.
\item Faceness-Net-SR-TP - a variant of the Faceness-Net-SR the uses the template proposal technique to generate candidate windows.
\end{itemize} 
The discussion on generic object proposal and template proposal techniques can be found in Sec.~\ref{subsec:region_proposal}.

\section{Results}
\label{sec:experiments}
\subsection{Evaluating the Attribute-Aware Networks}
\label{sec:evaluate_part_localisation}

\begin{figure}[b]
\begin{center}
\vskip -0.3cm
{\includegraphics[width=\linewidth]{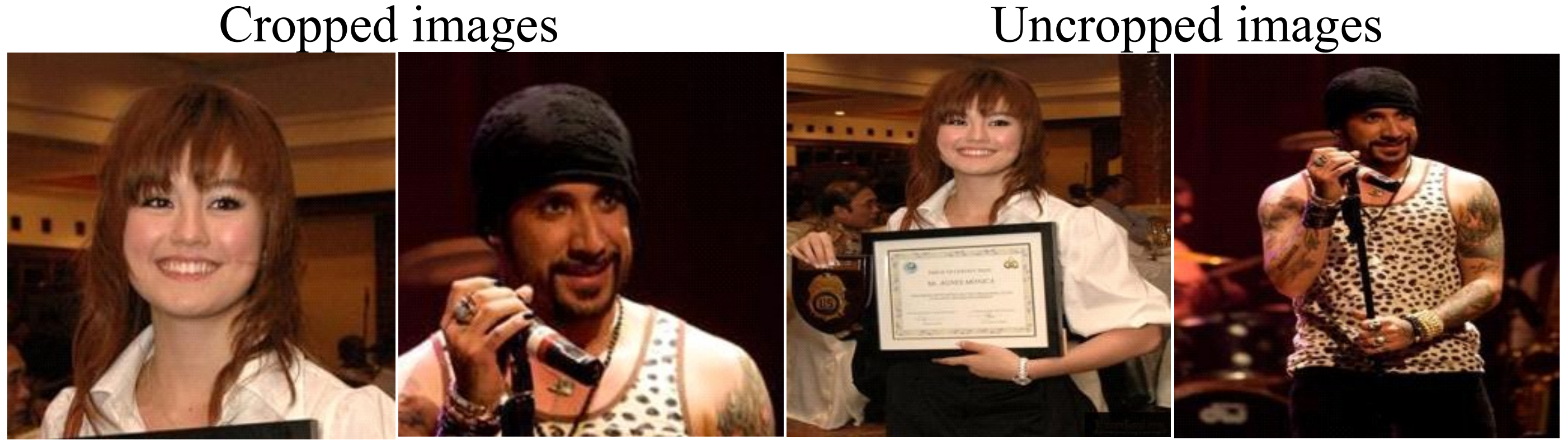}}
\vskip -0.25cm
\caption{Examples of cropped and uncropped images.}
\label{fig:cropped_uncropped_samples}
\end{center}
\end{figure}

\begin{table}[b]
\begin{center}
\caption{Evaluating the robustness to unconstrained training input. Facial part detection rate is used. The number of proposals is 350.}
\label{tab:acc_face_localization}
\scriptsize
\addtolength{\tabcolsep}{-1pt}
\begin{tabular}{c|c|c|c|c}
\hline
\textbf{Training Data} & \textbf{Hair}& \textbf{Eye}& \textbf{Nose}& \textbf{Mouth}
\\
\hline\hline
Cropped& $95.56\%$& $95.87\%$& $92.09\%$& $94.17\%$\\
Uncropped& $94.57\%$& $97.19\%$& $91.25\%$& $93.55\%$\\
\hline
\end{tabular}
\end{center}
\end{table}

\noindent \textbf{Robustness to unconstrained training input.}
The proposed attribute-aware networks do not assume well-cropped faces as input in both the training and test stages. 
To support this statement, we conduct an experiment by fine-tuning two attribute-aware networks as shown in Fig.~\ref{fig:attribute_aware_network}(a), each of which taking different inputs: (1) cropped images, which encompass roughly the face and shoulder regions, and (2) uncropped images, which may include large portions of background apart the face. Some examples of cropped and uncropped images are shown in Fig.~\ref{fig:cropped_uncropped_samples}.

The performances of these two networks are measured based on the part detection rate.
Note that we combine the evaluation on `Hair+Beard' to suit the ground truth provided by~\cite{crf} (see Sec.~\ref{sec:settings}).
We provide more details of part detection as follows. For each facial part, a total of five region templates are first defined using statistics obtained from the LFW training set. Non Maximum Suppression (NMS) is used to find the pixel locations with local maximum responses. We select the top $70$ NMS points and propose region templates centered at the points.

\begin{figure*}[t]
\begin{center}
\vspace{-0.5cm}
\includegraphics[width=0.8\linewidth]{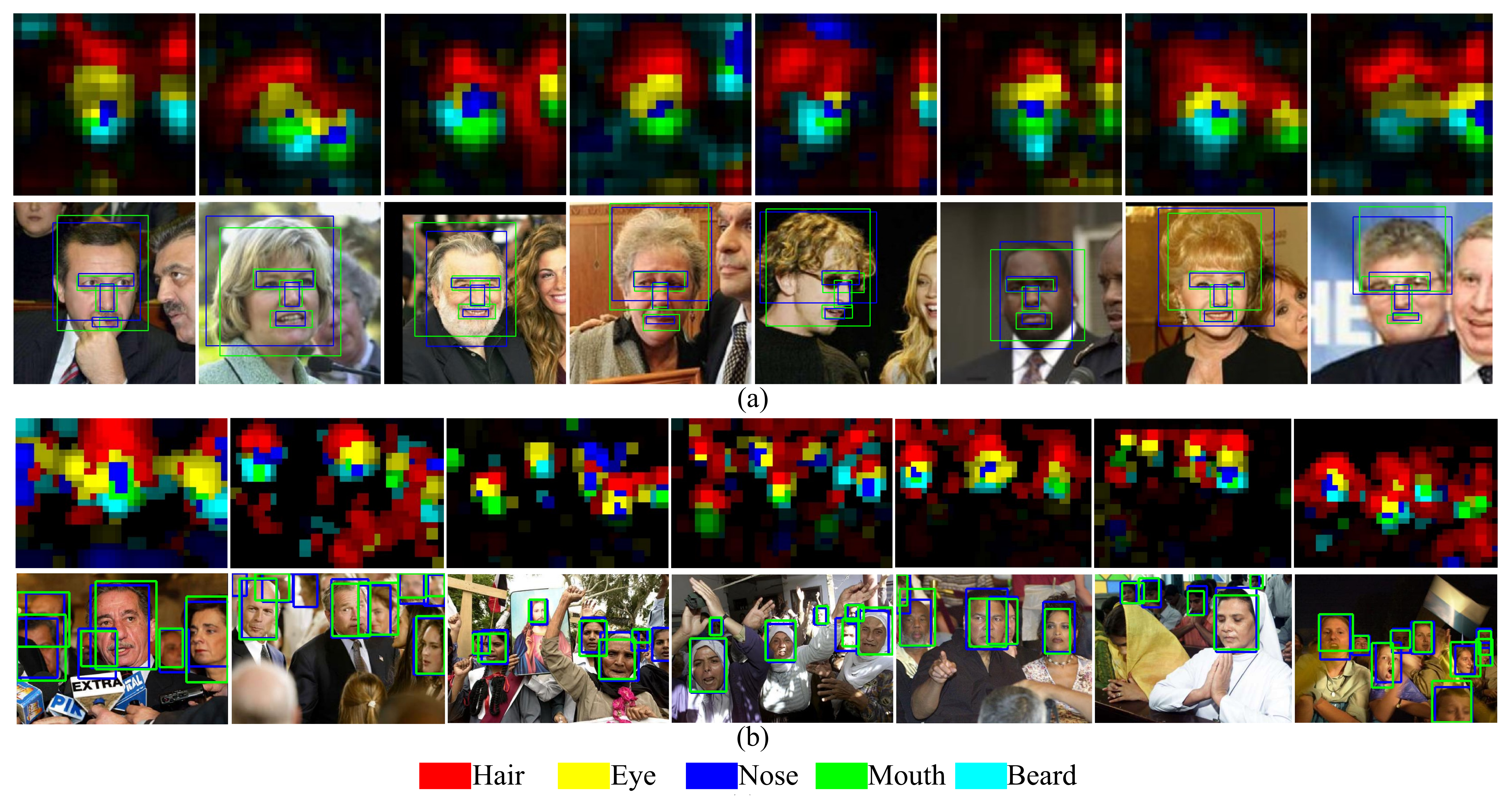}
\vskip -0.25cm
 {\caption{(a) The first row depicts the response maps generated by the proposed approach on each part. The second row shows the part localization results. Ground truth is depicted by the blue bounding boxes, while our part proposals are indicated in green.
(b) Face detection results on FDDB images. The bounding box in green is detected by our method while ground truth is printed in blue. We show the partness maps as reference. The results shown in (a) and (b) are generated using the Faceness-Net.
}
\vspace{-0.6cm}
\label{fig:vs_ex2}}
\end{center}
\end{figure*}

The detection results are summarized in Table~\ref{tab:acc_face_localization}.
As can be observed, the proposed approach performs similarly given both the cropped and uncropped images as training inputs.
The results suggest the robustness of the method in handling unconstrained images for training. In particular, thanks to the facial attribute-driven training, despite the use of uncropped images, the deep model is encouraged to discover and capture the facial part representation in the deep layers, it is therefore capable of generating response maps that precisely pinpoint the locations of parts.
The top row Fig.~\ref{fig:vs_ex2}(a) shows the partness maps generated from LFW images. The bottom row of Fig.~\ref{fig:vs_ex2}(a) shows the proposals which have the maximum overlap with the ground truth bounding boxes. Note that facial parts can be discovered despite challenging poses.
In the following experiments, all the proposed models are trained on uncropped images.

\begin{figure}[t]
\begin{center}
\includegraphics[width=\linewidth]{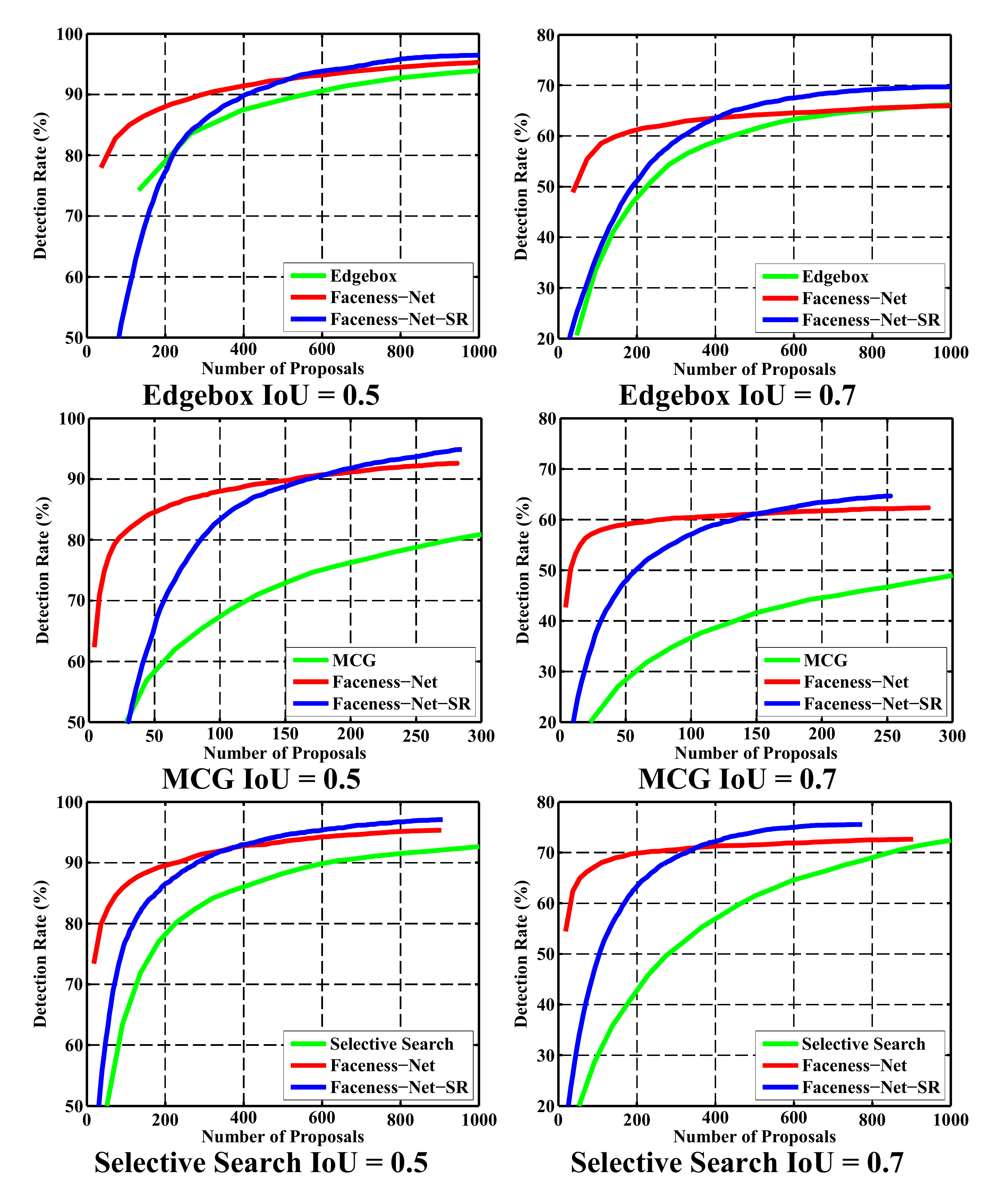}
\vskip -0.4cm
\caption{\small{We compare the performance between Faceness-Net, Faceness-Net-SR, and various generic objectness measures on proposing face candidate windows.}}
\label{fig:compare_generic_proposal}
\vspace{-0.45cm}
\end{center}
\end{figure}

\noindent \textbf{With and without sharing representation.}
As mentioned in Sec.~\ref{subsec:attribute_aware_networks}, we can train an attribute-aware network for each face part or we can train a single network for all the parts by sharing representation.
We compare the proposal detection rate of these two options.
Figure~\ref{fig:compare_generic_proposal} shows the proposal detection rate of the attribute-aware network(s) trained with and without sharing representation, indicated by blue and red curves, respectively.   
Attribute-aware networks trained without sharing representation require a fewer number of proposals but with a detection rate typically lower than $90\%$ (given 150-200 proposals).
On the contrary, the attribute-aware network that shares low-level and mid-level representations can achieve a higher detection rate but with an expense of a larger number of proposals.

The observations can be explained as follows. The networks without sharing representation tend to model the discrepancies between individual parts and background, while the network that shares representation is more likely to learn the differences between facial parts.
Thus, the latter has poorer background modelling capacity thus leading to inferior performance when the number of proposals is small, in comparison to the former. Nevertheless, we found that the network that shares representation yields high responses for subtle facial parts. This high recall rate is essential to improve the performance of face detection in the later stage.

\begin{figure}[t]
\begin{center}
\includegraphics[width=0.7\linewidth]{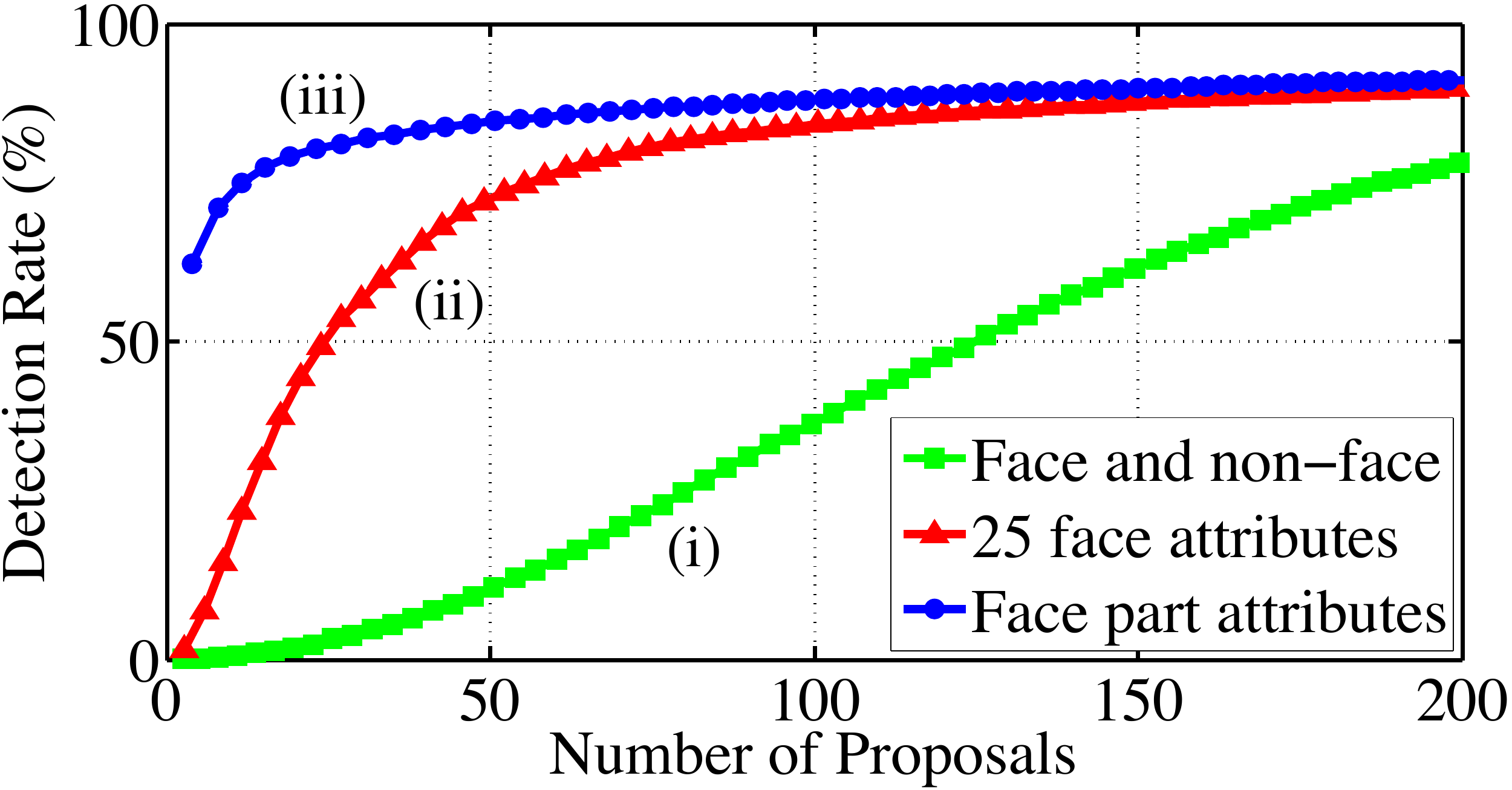}
\vskip -0.4cm
\caption{Comparing the face proposal performance when different strategies are used to fine-tune the attribute-aware networks.}
\label{fig:training_strategies}
\vspace{-0.45cm}
\end{center}
\end{figure}

\noindent
\textbf{Different fine-tuning strategies.}
As discussed in Sec.~\ref{subsec:learning_partness}, there are different fine-tuning strategies that can be considered for learning to generate a partness map, but not all of them are well-suited for deriving a robust faceness measure.
Qualitative results have been provided in Fig.~\ref{fig:partness_map_learning}. Here, we provide quantitative comparisons of face proposal performance between the following fine-tuning approaches: (i) a network fine-tuned with a large number of face images from CelebA and non-face images, (ii) fine-tuning the network with 25 face attributes, and (iii) the proposed approach that fine-tunes attribute-aware networks with part-level attributes in accordance to Table~\ref{tab:attribute_category}. It is evident from Fig.~\ref{fig:training_strategies} that our approach performs significantly better than approaches (i) and (ii).

\subsection{From Part Responses to Face Proposal}
\label{sec:evaluate_face_proposal}

\noindent \textbf{Generic object proposal methods.}
In this experiment, we show the effectiveness of adapting different generic object proposal generators~\cite{arbelaez2014multiscale, zitnick2014edge, uijlings2013selective} to produce face-specific proposals. Since the notion of face proposal is new, no suitable methods are comparable therefore we use the original generic methods as baselines.
We first apply any object proposal generator to generate the candidate windows and we use our faceness scoring method described in Sec.~\ref{subsec:faceness_measure} to obtain the face proposals.
We experiment with different parameters for the generic methods, and choose parameters that produce a moderate number of proposals with a very high recall.
Evaluation is conducted following the standard protocol~\cite{zitnick2014edge}.

The results are shown in Fig.~\ref{fig:compare_generic_proposal}. The green curves show the performance of baseline generic object proposal generators. It can be observed that our methods, both Faceness-Net and its variant Faceness-Net-SR, consistently improve the baselines for proposing face candidate windows, under different IoU thresholds. Table~\ref{tab:compare_generic_proposal} shows that our method achieves high detection rate with moderate number of proposals.

\begin{table}[t]
\begin{center}
\caption{The number of proposals needed for different detection rate.}
\label{tab:compare_generic_proposal}
\vspace{-0.3cm}
\scriptsize
\addtolength{\tabcolsep}{-1pt}
\begin{tabular}{c|c|c|c|c}
\hline
\textbf{Proposal method} & \textbf{75\%}& \textbf{80\%}& \textbf{85\%}& \textbf{90\%}
\\
\hline\hline
EdgeBox~\cite{zitnick2014edge}& $132$& $214$& $326$& $600$\\
EdgeBox~\cite{zitnick2014edge}+Faceness& $\mathbf{21}$& $\mathbf{47}$& $\mathbf{99}$& $\mathbf{288}$ \\
EdgeBox~\cite{zitnick2014edge}+Faceness-SR& $180$& $210$& $275$& $380$ \\
\hline
MCG~\cite{arbelaez2014multiscale}& $191$& $292$& $453$& $942$\\
MCG~\cite{arbelaez2014multiscale}+Faceness& $\mathbf{13}$& $\mathbf{23}$& $\mathbf{55}$& $\mathbf{158}$ \\
MCG~\cite{arbelaez2014multiscale}+Faceness-SR& $69$& $87$& $112$& $158$ \\
\hline
Selective Search~\cite{uijlings2013selective}& $153$& $228$& $366$& $641$ \\
Selective Search~\cite{uijlings2013selective}+Faceness& $\mathbf{24}$& $\mathbf{41}$& $\mathbf{91}$& $\mathbf{237}$ \\
Selective Search~\cite{uijlings2013selective}+Faceness-SR& $94$& $125$& $189$& $309$ \\
\hline
EdgeBox~\cite{zitnick2014edge}+Faceness& $\mathbf{21}$& $\mathbf{47}$& $\mathbf{99}$& $\mathbf{288}$ \\
Template Proposal+Faceness& $183$& $226$& $256$& $351$ \\
Template Proposal+Faceness(without re-scored)& $447$& $517$& $621$& $847$ \\
\hline
\end{tabular}
\vspace{-0.45cm}
\end{center}
\end{table}

\begin{figure}[t]
\begin{center}
\includegraphics[width=\linewidth]{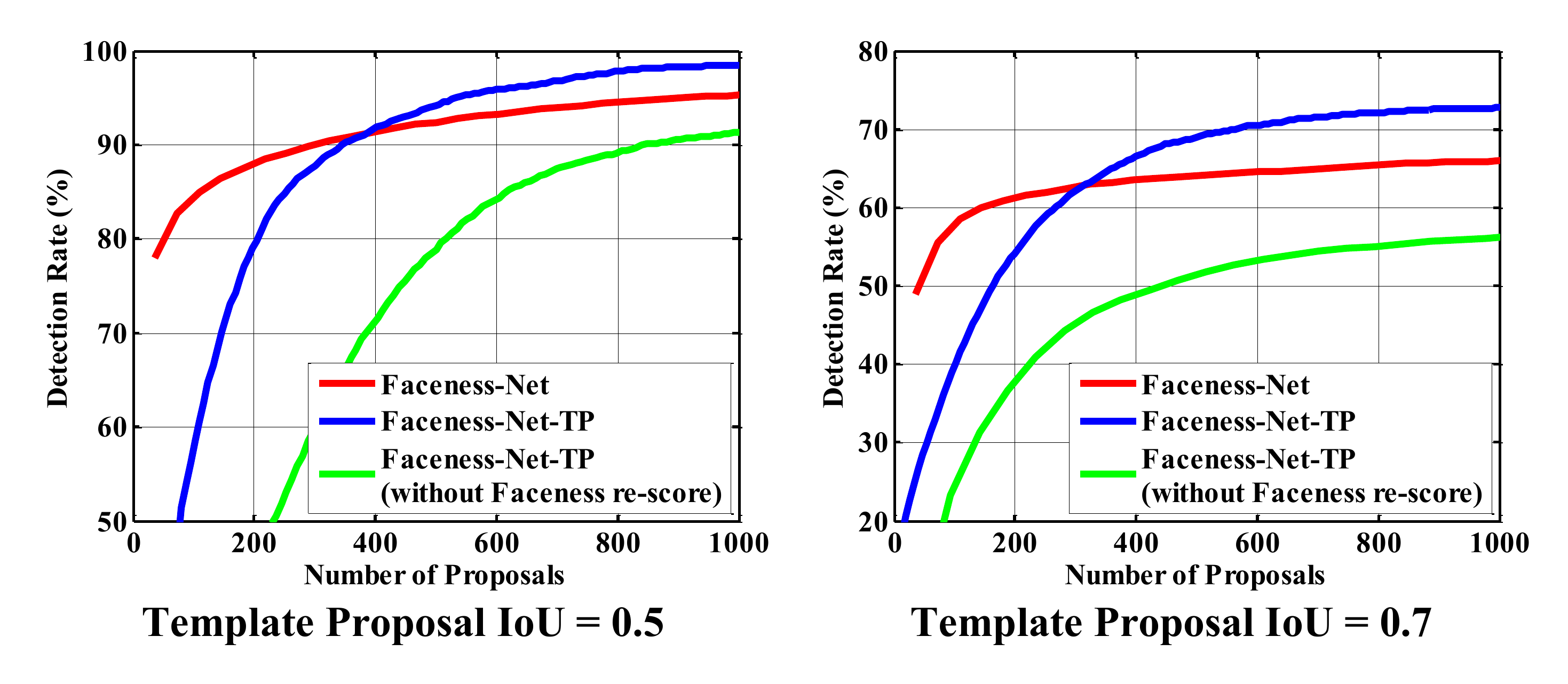}
\vskip -0.4cm
\caption{\small{We compare the performance between Faceness-Net and Faceness-Net-TP. The former uses a generic object proposal method for proposing face candidate windows, while the latter employs the template proposal method presented in Section~\ref{subsec:region_proposal}. We also compared against a baseline Faceness-Net-TP (without Faceness re-score) to show the importance of rescoring the candidate windows with Faceness score.}}
\label{fig:compare_region_template_proposal}
\vspace{-0.45cm}
\end{center}
\end{figure}

\noindent \textbf{Template proposal method.}
In this experiment, we compare face proposal performance by using three different methods for generating and scoring candidate windows: 
\begin{enumerate}
\item The original Faceness-Net in which an external generic object proposal generator is adopted for generating candidate windows. The candidate windows are re-ranked using the faceness score. The result is shown in Fig.~\ref{fig:compare_region_template_proposal} indicated with a red curve.
\item A variant of the Faceness-Net, named as Faceness-Net-TP, that adopts the template proposal technique (Section~\ref{subsec:region_proposal}) to generate candidate windows. The candidate windows are re-ranked using the faceness score. The result is shown in Fig.~\ref{fig:compare_region_template_proposal} indicated with a blue curve.
\item The baseline is a Faceness-Net-TP, of which candidate windows are not re-ranked using the faceness score. Specifically, given a normalized partness map, we find pixel locations where response values are equal or higher than a threshold. Then, templates are applied centered at selected locations to generate template proposals without using faceness score to re-scoring and re-ranking proposals. The result is shown in Fig.~\ref{fig:compare_region_template_proposal} indicated with a green curve.
\end{enumerate}

One can observe from Fig.~\ref{fig:compare_region_template_proposal} that Faceness-Net outperforms Faceness-Net-TP when the number of proposals is fewer than $300$ (the low-recall region). 
The performance gap is likely caused by the quality of initial candidate windows generated by the generic object proposal (used by Faceness-Net) and template proposal (used by Faceness-Net-TP). The former, such as EdgeBox, employs Structured Edges as informative representation to generate the initial set of candidate windows. The latter, on the other hand, starts with a set of pre-determined template boxes. 
Despite the lower performance at low-recall region, Faceness-Net-TP achieves a high detection rate of over $96.5\%$ at high-recall region, which is not achievable using Faceness-Net that employs generic object proposal.
Moreover, the computation cost of template proposal is much lower than generic object proposal.

\begin{figure}[t]
\begin{center}
\includegraphics[width=0.7\linewidth]{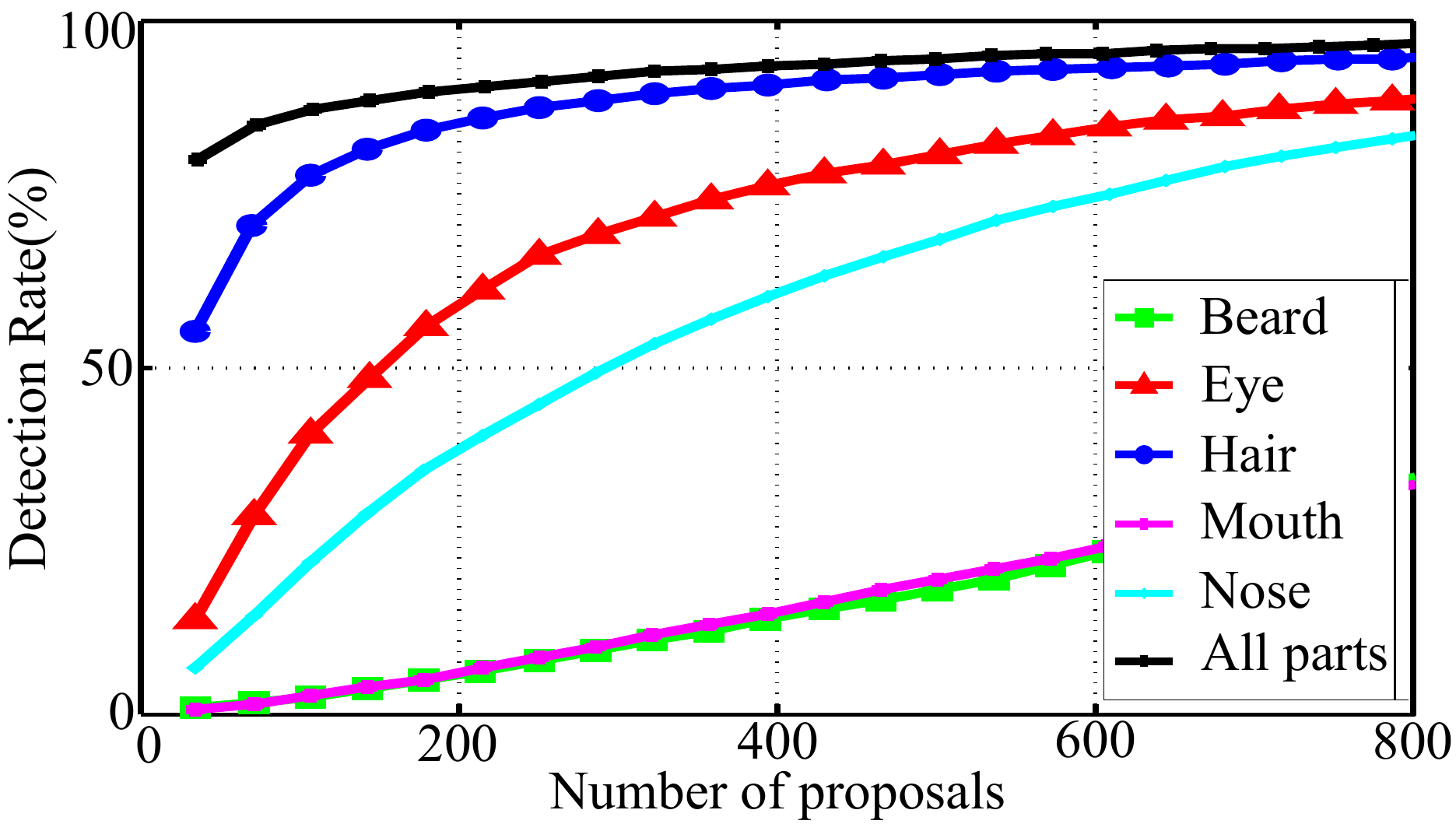}
\vskip -0.4cm
\caption{The contributions of different face parts to face proposal.}
\label{fig:part_contribution}
\vspace{-0.45cm}
\end{center}
\end{figure}

\noindent \textbf{Evaluate the contributions of each face part.}
We factor the contributions of different face parts to proposing face.
Specifically, we generate face proposals with partness maps from each face part individually using the same evaluation protocol in previous experiment.
As can be observed from Fig.~\ref{fig:part_contribution}, the hair, eye, and nose parts perform much better than mouth and beard.
The lower part of the face is often occluded, making the mouth and beard less effective in proposing face windows.
In contrast, hair, eye, and nose are visible in most cases. They therefore become important clues for face proposal. Nonetheless, mouth and beard could provide complementary cues. Thus combining all parts leads to better result than considering each part in isolation.

\subsection{From Face Proposal to Face Detection}
\label{sec:evaluate_face_detection}

Next, we compare the proposed Faceness-Net and its variants against state-of-the-art face detection approaches on four benchmark datasets FDDB~\cite{fddb}, AFW~\cite{zhu2012face}, PASCAL faces~\cite{yan2014face} and WIDER FACE~\cite{yang2016wider}.
Our baseline face detector, Faceness-Net, which involves five CNNs with the structure shown in the Fig.~\ref{fig:baseline_facenessnet}, is trained with the top $200$ proposals by re-ranking MCG proposals following the process described in Sec.~\ref{subsec:faceness_measure}.
To factor the contributions of share representation and template proposal, we build another three variants of Faceness-Net as discussed in Sec.~\ref{sec:settings}. 
The variant Faceness-Net-TP is trained with the top $1000$ template proposals that are re-ranked following Sec.~\ref{subsec:faceness_measure}.

We compare Faceness-Net and its variants against representative published methods~\cite{ACF-multiscale,HeadHunter,JointCascade,FastDPM,BoostedExemplar,SURF,PEP-Adapt,XZJY,zhu2012face,VJain} on FDDB.
For the PASCAL faces and AFW we compare with (1) deformable part based methods, \eg~ structure model~\cite{yan2014face} and Tree Parts Model (TSM)~\cite{zhu2012face}; (2) cascade-based methods, \eg,~Headhunter~\cite{HeadHunter}.
For the WIDER FACE~\cite{yang2016wider} we compare with (1) aggregated channel feature method (ACF)~\cite{ACF-multiscale}; (2) deformable part based model~\cite{HeadHunter}; (3) cascaded-based method~\cite{VJain}.

\begin{figure}[t]
\begin{center}
\includegraphics[width=0.85\linewidth]{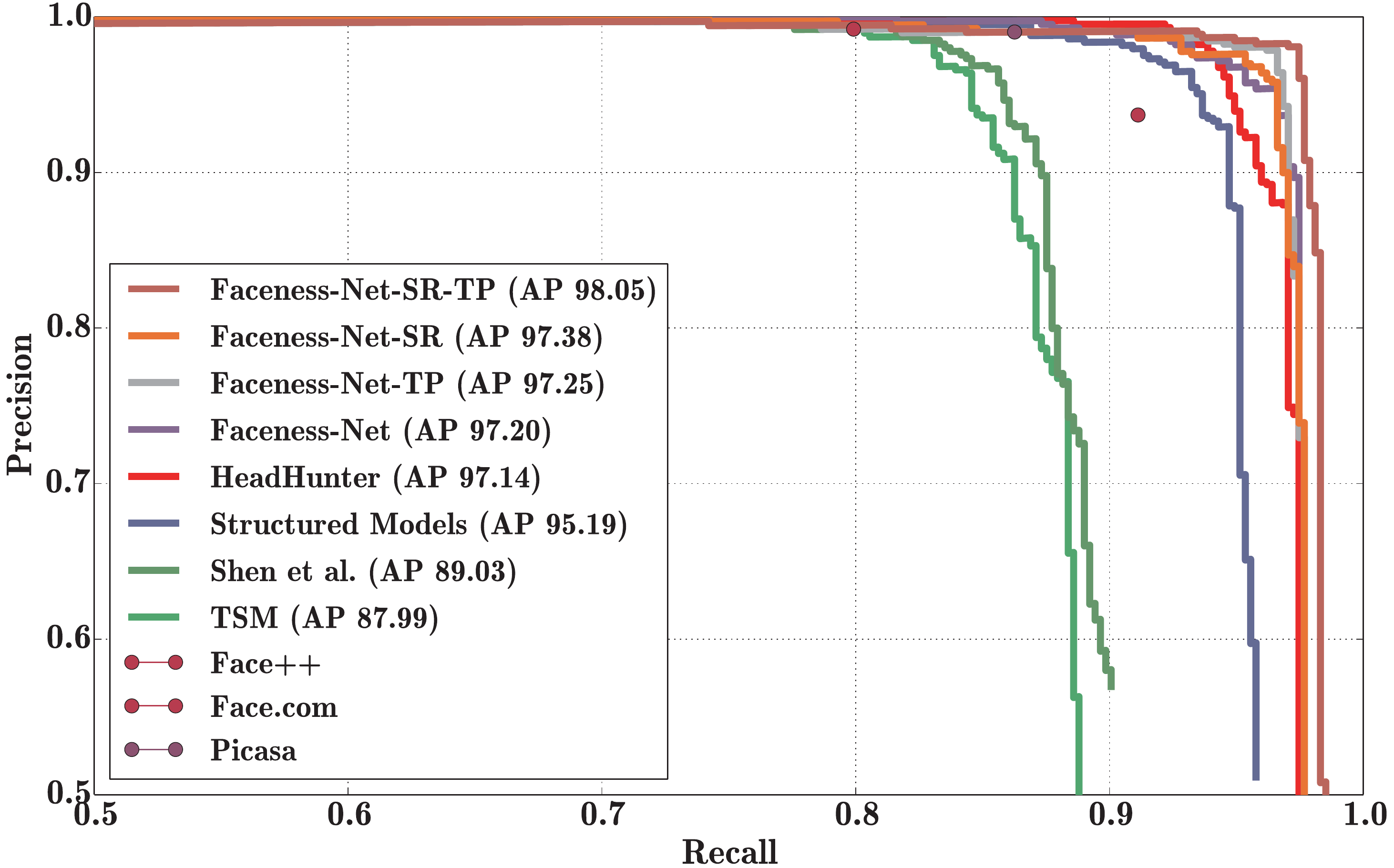}
\vskip -0.4cm
\caption{Precision-recall curves on the AFW dataset. AP = average precision.}
\label{fig:face_detection_afw}
\vspace{-0.45cm}
\end{center}
\end{figure}

\begin{figure}[t]
\begin{center}
\includegraphics[width=0.85\linewidth]{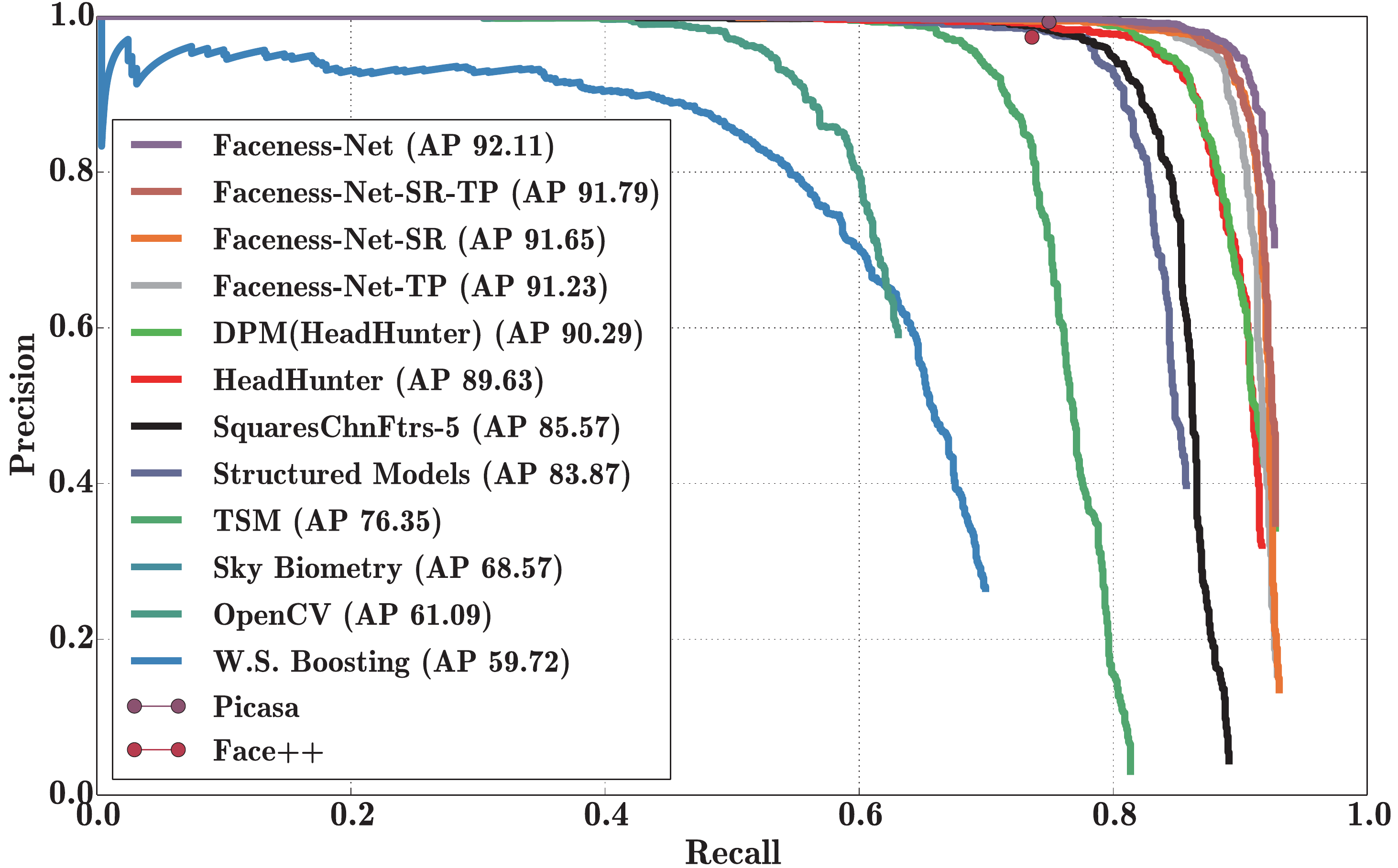}
\vskip -0.4cm
\caption{Precision-recall curves on the PASCAL faces dataset. AP = average precision.}
\label{fig:face_detection_pascal}
\vspace{-0.45cm}
\end{center}
\end{figure}

\begin{figure}[t]
\begin{center}
\includegraphics[width=\linewidth]{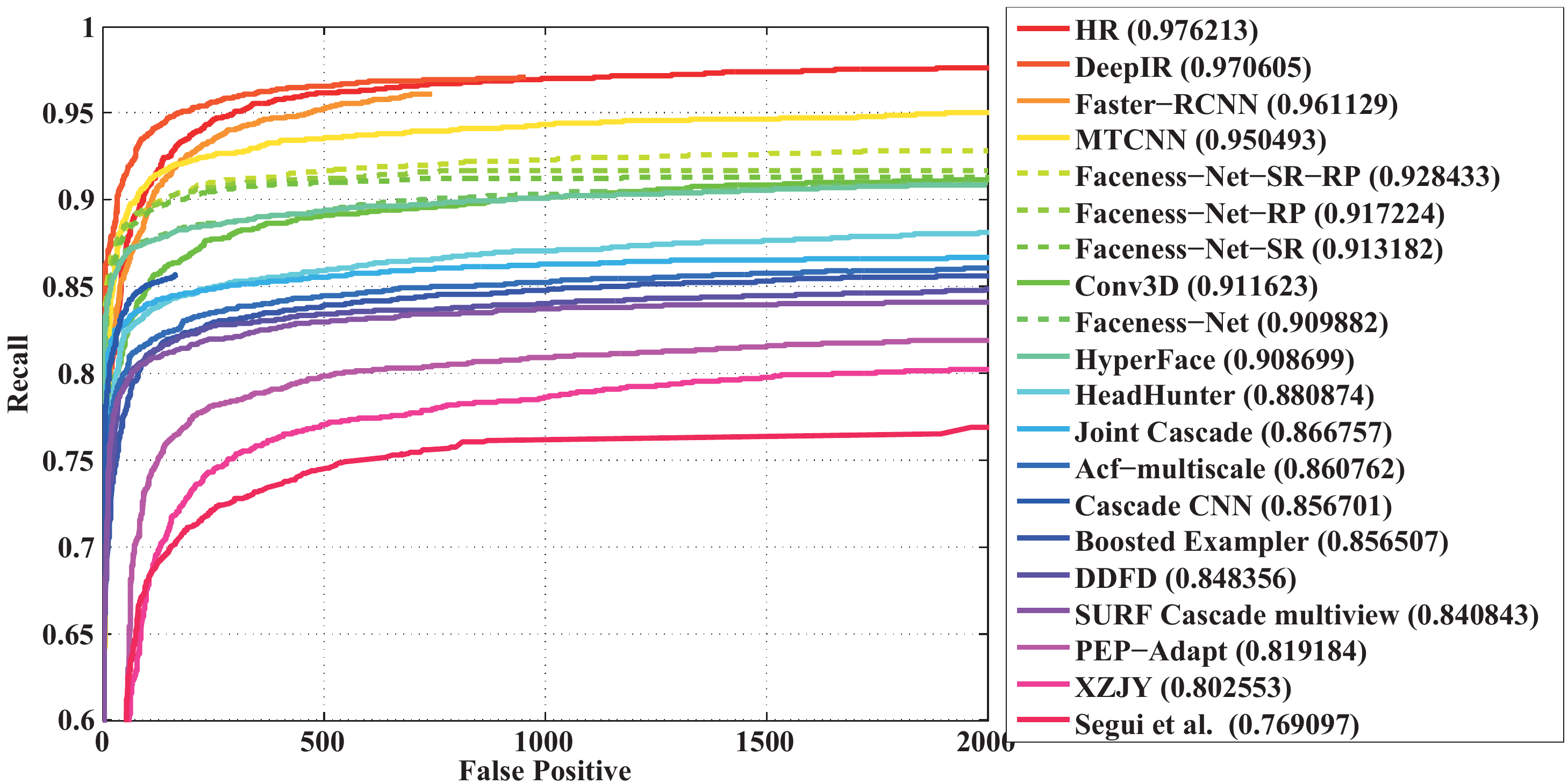}
\vskip -0.4cm
\caption{FDDB results evaluated using discrete sore.}
\label{fig:face_detection_fddb_dist}
\vspace{-0.45cm}
\end{center}
\end{figure}

\begin{figure}[t]
\begin{center}
\includegraphics[width=\linewidth]{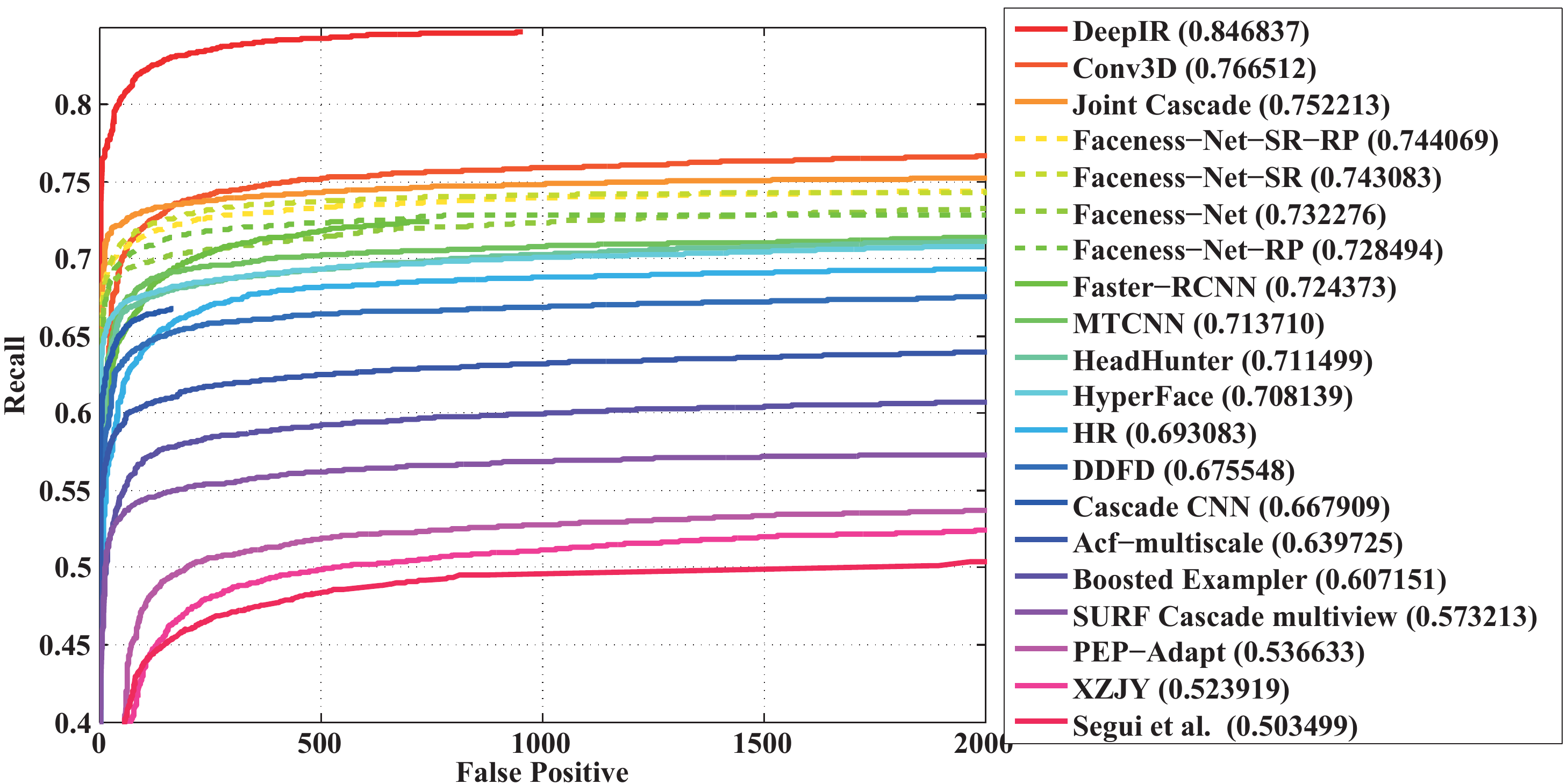}
\vskip -0.4cm
\caption{FDDB results evaluated using continuous sore.}
\label{fig:face_detection_fddb_cont}
\vspace{-0.45cm}
\end{center}
\end{figure}

\noindent \textbf{AFW dataset.}
Figures~\ref{fig:face_detection_afw} shows the precision and recall curves of the compared face detection algorithms on the AFW dataset.
We observe that Faceness-Net and its variants outperform all the compared approaches by a considerable margin. 
The Faceness-Net-SR and Faceness-Net-TP outperform baseline Faceness-Net, suggesting the effectiveness of sharing representation and template proposal technique.
Among all the Faceness-Net variants, Faceness-Net-SP-TP achieves the best performance with a high average precision of $98.05\%$.

\noindent \textbf{PASCAL faces dataset.}
Figure~\ref{fig:face_detection_pascal} shows the precision and recall curves. The baseline Faceness-Net outperforms its variants and other compared face detection algorithms.
Compared with other benchmark datasets, PASCAL faces dataset has a fewer number of faces in each image, therefore only a small number of proposals is required to achieve a high recall rate.  
As shown in Fig.~\ref{fig:compare_generic_proposal}, the quality of proposals generated by the baseline Faceness-Net is higher than its variants when the number of proposals is lower than $200$, which leads to its better face detection performance on PASCAL face dataset.

\noindent \textbf{FDDB dataset.}
The results are shown in Fig.~\ref{fig:face_detection_fddb_dist} and Fig.~\ref{fig:face_detection_fddb_cont}.
Faceness-Net and its variants achieve competitive performance compared with existing algorithms evaluated using the discrete score as shown in the Fig.~\ref{fig:face_detection_fddb_dist}.
Faceness-Net baseline achieves $90.99\%$ recall rate, while Faceness-Net-SR and Faceness-Net-TP outperform the baseline Faceness-Net by $0.4\%$ and $0.7\%$, respectively.
Faceness-Net-SR-TP performs best with a large improvement of $1.85\%$ compared with the baseline Faceness-Net.

\begin{figure*}[t]
\begin{center}
\includegraphics[width=\linewidth]{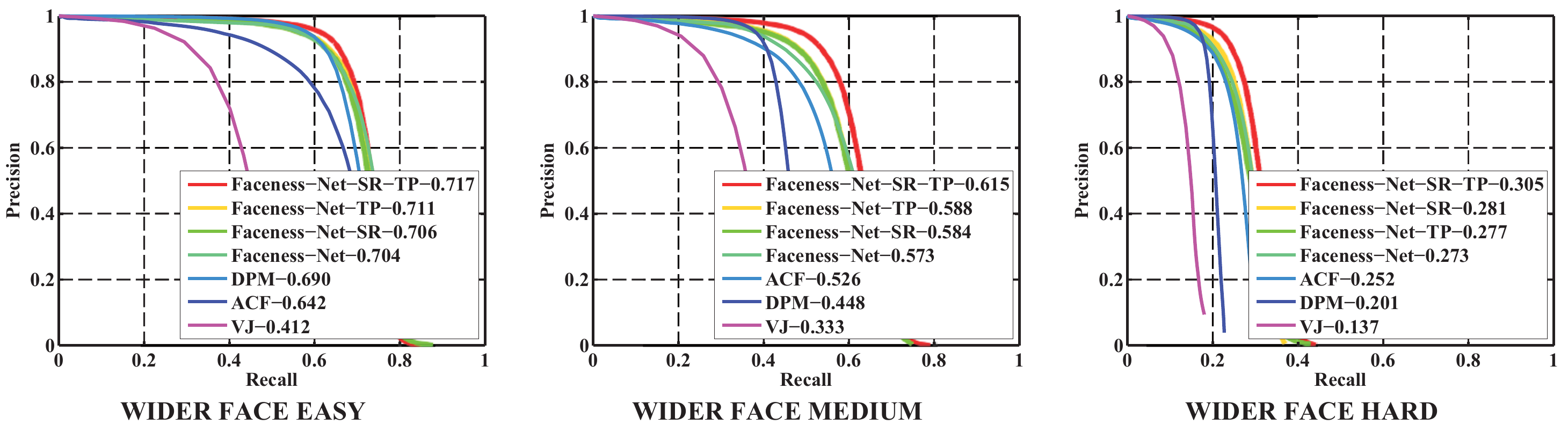}
 {\caption{Precision and recall curves of different subsets of WIDER FACES: Overall Easy/Medium/Hard subsets. AP = average precision.}
\label{fig:face_detection_wider}}
\vspace{-0.45cm}
\end{center}
\end{figure*}

\noindent \textbf{WIDER FACE dataset.}
WIDER FACE dataset is currently the largest face detection benchmark dataset. The dataset has two evaluation protocols. The internal protocol evaluates face detection algorithms that use WIDER FACE data during training. In contrast, the external protocol evaluates face detection algorithms that are trained on external data.
Since Faceness-Net and its variants are trained on CelebA and AFLW datasets without using images in the WIDER FACE dataset, we evaluate our algorithm using the external setting.
Faceness-Net and its variants yield better performance in all three evaluation settings compared with baseline method, namely ``Easy'', ``Medium'', and ``Hard'' as shown in Fig.~\ref{fig:face_detection_wider}.
The variants of Faceness-Net outperform baseline Faceness-Net by a considerable margin, suggesting the effectiveness of representation sharing and template proposal techniques.
Although Faceness-Net and its variants outperform baseline methods under external setting, there exist large gap between Faceness-Net and recent state-of-the-art methods~\cite{zhang2016joint,zhu2016cms,hu2016finding,yang2017face}. 
These methods are trained using the WIDER FACE dataset and thus they can deal with more challenging cases. On the contrary, our method is trained on datasets that are not targeted for face detection (CelebA for face recognition, and AFLW for face alignment) and with simple backgrounds. Nevertheless, it still achieves promising performance. We provide more discussion below.

\begin{table}[t]
\begin{center}
\caption{A comparison of training data and annotations adopted in state-of-the-art face detection methods.}
\vskip -0.35cm
\label{tab:cmp_training_data}
\tiny
\addtolength{\tabcolsep}{-1pt}
\begin{tabular}{c|c|c|c|c|c|c|c}
\hline
& \rotatebox{90}{\textbf{Dataset}}& \rotatebox{90}{\textbf{\#Images}} & \rotatebox{90}{\textbf{\#Bounding Boxes}}& \rotatebox{90}{\textbf{\#Landmarks}}& \rotatebox{90}{\textbf{\#Attribute}}& \rotatebox{90}{\textbf{Clutter Background}}& \rotatebox{90}{\textbf{ImageNet pretrain}}
\\
\hline\hline
Faceness-Net& CelebA+AFLW & $180$k+$13$k & $17$k & - & $26$ & - & -\\
STN~\cite{chen2016supervised}& Internal Dataset+MS COCO & $400$k+$120$k & $<400$k & $5$ & - & \Checkmark & - \\
Faster-RCNN~\cite{jiang2016face}& WIDER FACE & $13$k & $150$k & - & - & \Checkmark & \Checkmark\\
MTCNN~\cite{zhang2016joint}& CelebA+WIDER FACE & $200$k+$13$k & $350$k & $5$ & - & \Checkmark & - \\
\hline
\end{tabular}
\end{center}
\vspace{-0.4cm}
\end{table} 

\begin{figure}[t]
\begin{center}
\includegraphics[width=0.9\linewidth]{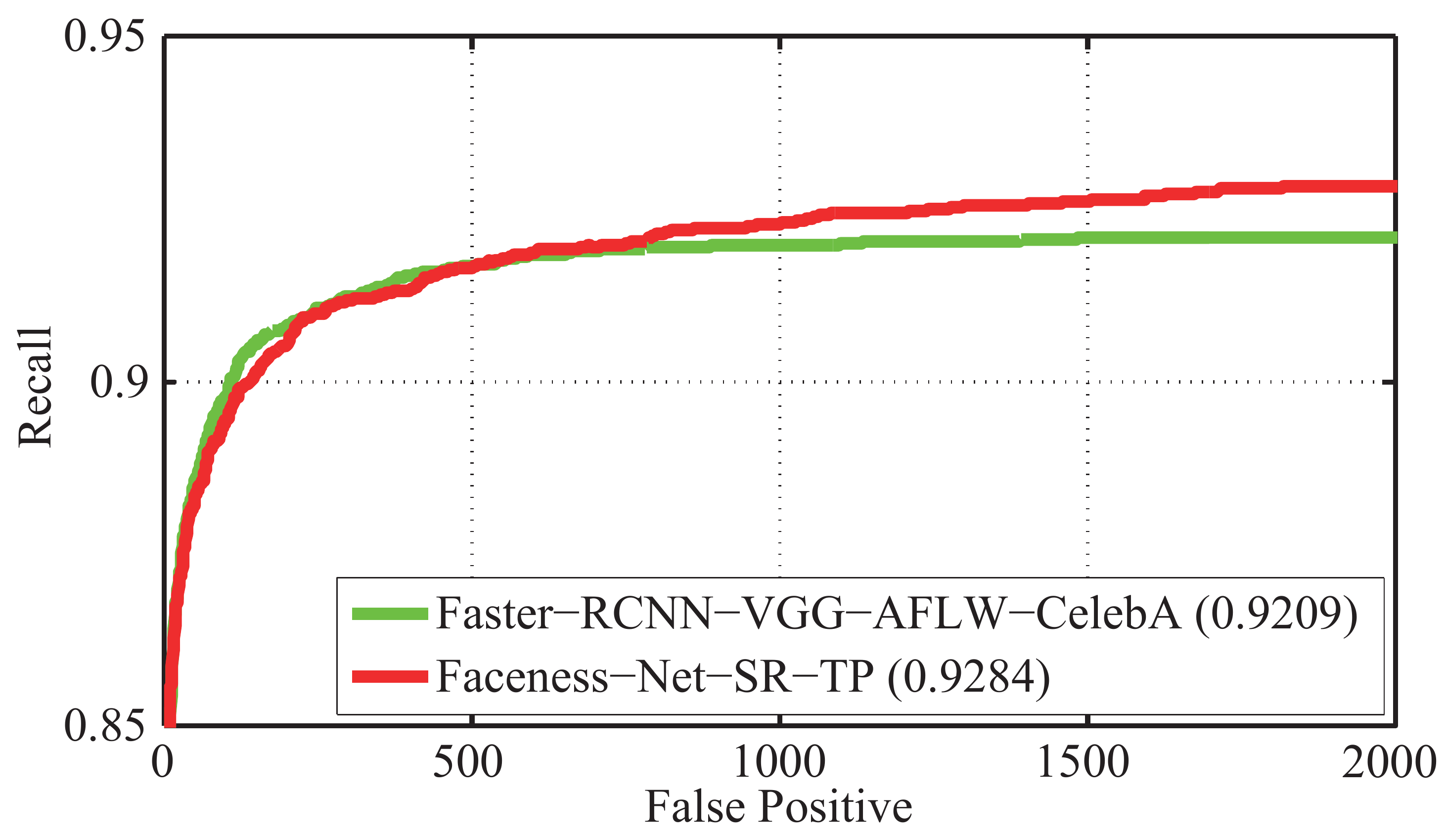}
\caption{A comparison of Faster-RCNN face detector~\cite{jiang2016face} and Faceness-Net on FDDB, when both methods adopt the same training data.}
\vskip -0.35cm
\label{fig:quantitative_training_data}
\vspace{-0.3cm}
\end{center}
\end{figure}

\noindent \textbf{Discussion:}
Recent studies~\cite{chen2016supervised,jiang2016face,zhang2016joint,yang2017face} achieve better face detection performance on FDDB, AFW, and PASCAL faces datasets compared to our Faceness-Net.
The performance gap between Faceness-Net and other methods arises from two aspects, namely, the better modeling of background clutter and stronger supervision signals.
Table~\ref{tab:cmp_training_data} summarizes the training data and supervision signals used by different algorithms.
Faceness-Net is trained on CelebA and AFLW datasets. These datasets are originally proposed for face recognition and facial landmark detection, respectively. The background in CelebA and AFLW is less cluttered and diverse compared with various backgrounds available in WIDER FACE and MS-COCO datasets.
In addition, faces in CelebA and AFLW datasets have smaller variations, both in scale and poses, compared to those captured in the WIDER FACE dataset. We use $17$k face bounding boxes compared to more than $150$k face bounding boxes employed by other methods.

To gain a fairer comparison, we train the Faster-RCNN model presented in~\cite{jiang2016face} using the same training sets (AFLW and CelebA) employed by Faceness-Net. Evaluation is performed on the FDDB dataset.
The results are shown in Fig.~\ref{fig:quantitative_training_data}. The Faster-RCNN face detector achieves $92.09\%$ detection rate on the FDDB dataset which is marginally lower than that of Faceness-Net. Note that, Faceness-SR-TP is not finetuned by using ImageNet data, but still achieves better performance than Faster-RCNN. This is probably because attribute supervisions are more capable of modeling facial parts.

Apart from using more challenging training images, both STN~\cite{chen2016supervised} and MTCNN~\cite{zhang2016joint} use facial landmarks to localize face. Facial landmarks indicate the explicit location of face parts and thus provide stronger supervisory information than face attributes. 
Our method can benefit from these additional factors.  Specifically, it is possible to obtain a stronger Faceness-Net detector using facial landmarks based supervision and datasets with a more cluttered background.

\begin{figure*}[t]
\begin{center}
{\includegraphics[width=0.9\linewidth]{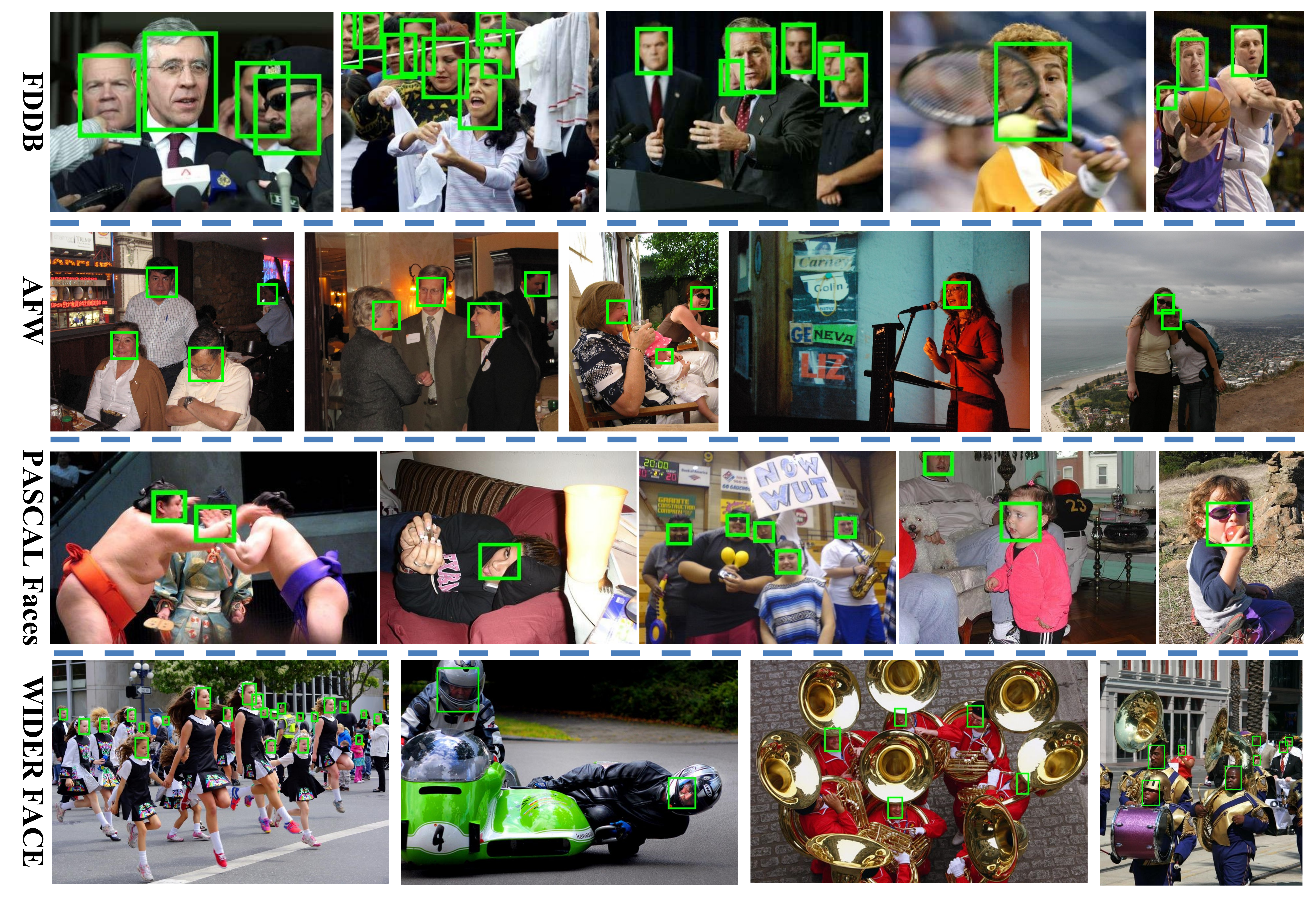}}
\vskip -0.35cm  
{\caption{Face detection results obtained by Faceness-Net on FDDB, AFW, PASCAL faces, and WIDER FACE.}\label{fig:qualitative_result}}
\vspace{-0.5cm}
\end{center}
\end{figure*}

\begin{figure*}[t]
\begin{center}
\includegraphics[width=0.9\linewidth]{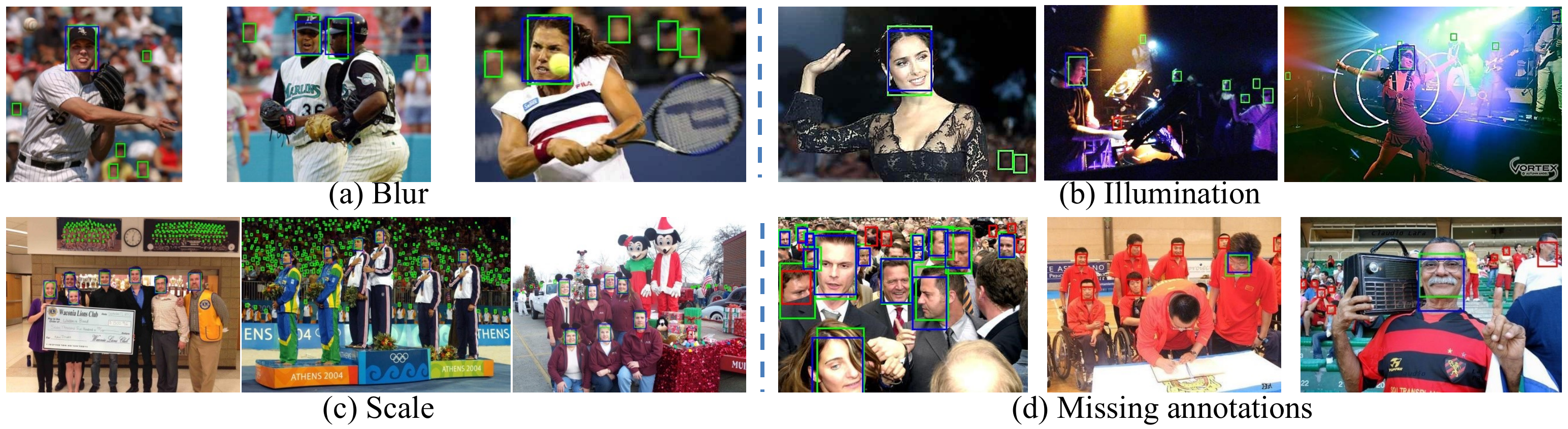}
\vskip -0.35cm
{\caption{Failure cases of Faceness-Net. The bounding box in green is ground truth. Our detection result is printed in blue. Bounding boxes in red indicate faces that are not annotated but detected by our detector (Best viewed in color).}
\label{fig:failure_case}}
\vspace{-0.5cm}
\end{center}
\end{figure*}

Finally, we show some qualitative examples in Fig.~\ref{fig:qualitative_result}.
Some failure cases are provided in Fig.~\ref{fig:failure_case}. The failures are mainly caused by blurring, illumination, tiny face scale, and missed annotations. 
Among the various causes, tiny faces (with a resolution as low as 20 pixels height) remain one of the hardest issues that we wish to further resolve.
The visual appearances between tiny and normal-size faces exhibit a huge difference. In particular, the facial parts such as eyes, nose or mouth can be barely distinguished from tiny faces, which makes responses produced by attribute-aware networks meaningless.
In order to recall tiny faces, data augmentation and multi-scale inference may be adopted. Nonetheless, learning scale-invariant representation is still an open problem.
In this study, we do not deal with tiny faces explicitly. It is part of our on-going work~\cite{yang2017face}.

\section{Runtime analysis}

The runtime of the proposed Faceness-Net-SR-TP is $40$ms on a single GPU\footnote{The runtime is measured on a Nvidia Titan X GPU.}.
The time includes $10$ms to generate faceness proposals with the height of testing image no more than $300$ pixels.
The efficiency of Faceness-Net-SR-TP is clearly faster than the baseline Faceness-Net since the former shares the layers from conv1 to conv4 in its attribute-aware networks. 
Previous CNN based face detector~\cite{cascadecnn} achieves good runtime efficiency too. Our method differs significantly to this method in that we explicitly handle partial occlusion by inferring face likeliness through part responses.
This difference leads to a significant margin of $4.66\%$ in recall rate (Cascade-CNN $85.67\%$, our method $90.33\%$) when the number of false positives is fixed at $167$ on the FDDB dataset.
The complete recall rate of the proposed Faceness-Net-SR-TP is $92.84\%$ compared to $85.67\%$ of Cascade-CNN.
At the expense of recall rate, the fast version of Cascade-CNN achieves $14$fps on CPU and $100$fps on GPU for $640\times480$ VGA images.
Our Faceness-Net-SR-TP can achieve practical runtime efficiency under the aggressive setting mentioned above, but still with a $0.21\%$ higher recall rate than the Cascade-CNN.

\section{Conclusion}
\label{sec:conclusion}

Different from existing face detection studies, we explored the usefulness of face attributes based supervision for learning a robust face detector. We observed an interesting phenomenon that face part detectors  can be obtained from a CNN that is trained on recognizing attributes from uncropped face images, without explicit part supervision.
Consequently, we introduced the notion of `faceness' score, which was carefully formulated through considering facial parts responses and the associated spatial arrangements.
The faceness score can be employed to re-rank candidate windows of any region proposal techniques to generate a modest set of high-quality face proposals with high recall.
With the generated face proposals, we trained a strong face detector that demonstrated promising performance on various face detection benchmark datasets.

\ifCLASSOPTIONcompsoc
  \section*{Acknowledgments}
\else
   regular IEEE prefers the singular form
  \section*{Acknowledgment}
\fi

This work is supported by SenseTime Group Limited, the Hong Kong Innovation and Technology Support Programme, and the General Research Fund sponsored by the Research Grants Council of the Hong Kong SAR (CUHK 416713, 14241716, 14224316. 14209217).

\ifCLASSOPTIONcaptionsoff
  \newpage
\fi



%

\bibliographystyle{IEEEtran}
\bibliography{long,faceness_final}

\begin{IEEEbiography}[{\includegraphics[width=1in,height=1.25in,clip,keepaspectratio]{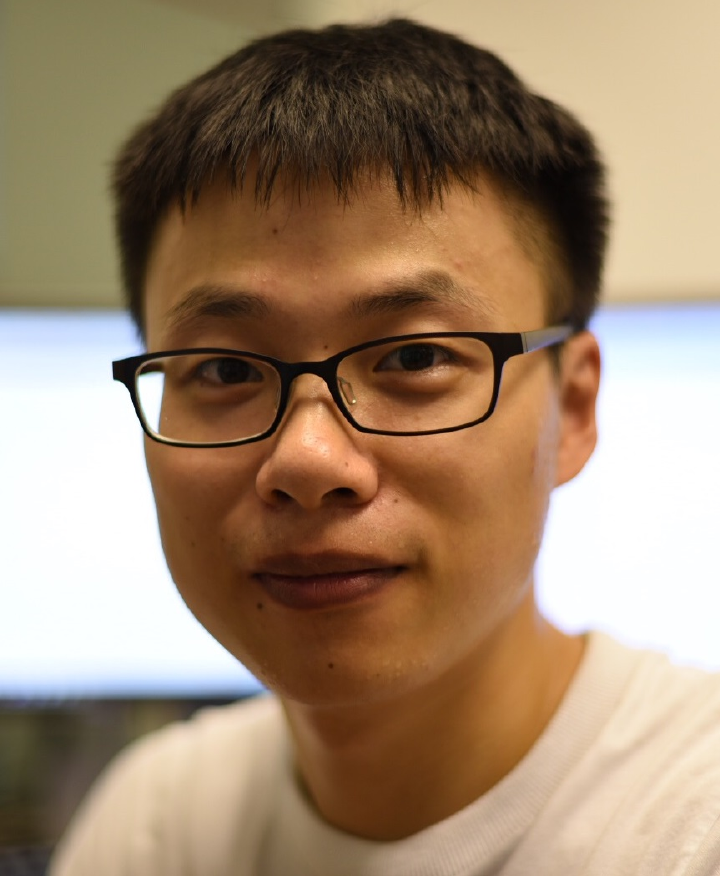}}]{Shuo Yang}
received the BE degree in software engineering from Wuhan University, China, in 2013. He is currently working toward the PhD degree in the Department of Information Engineering, Chinese University of Hong Kong. His research interests include computer vision and machine learning, in particular, face detection and object recognition. He is a student member of the IEEE.
\end{IEEEbiography}
\vspace{-0.7cm}
\begin{IEEEbiography}[{\includegraphics[width=1in,height=1.25in,clip,keepaspectratio]{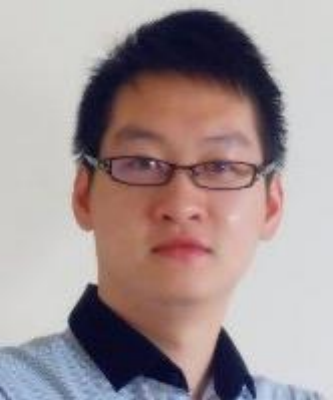}}]{Ping Luo}
received his PhD degree in 2014 in Information Engineering, Chinese University of Hong Kong (CUHK). He is currently a Research Assitant Professor in Electronic Engineering, CUHK. His research interests focus on deep learning and computer vision, including optimization, face recognition, web-scale image and video understanding. He has published 40+ peer-reviewed articles in top-tier conferences such as CVPR, ICML, NIPS and journals such as TPAMI and IJCV. He received a number of awards for his academic contribution, such as Microsoft Research Fellow Award in 2013 and Hong Kong PhD Fellow in 2011. He is a member of IEEE.
\end{IEEEbiography}
\vspace{-0.7cm}
\begin{IEEEbiography}[{\includegraphics[width=1in,height=1.25in,clip,keepaspectratio]{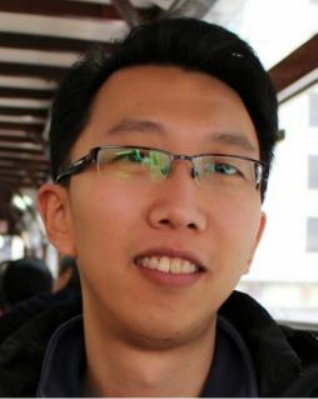}}]{Chen Change Loy}
received the PhD degree in computer science from the Queen Mary University of London in 2010. He is currently a research assistant professor in the Department of Information Engineering, Chinese University of Hong Kong. Previously, he was a postdoctoral researcher at Queen Mary University of London and Vision Semantics Ltd. He serves as an associate editor of IET Computer Vision Journal and a Guest Editor of Computer Vision and Image Understanding. His research interests include computer vision and pattern recognition, with focus on face analysis, deep learning, and visual surveillance. He is a senior member of the IEEE.
\end{IEEEbiography}
\vspace{-0.7cm}
\begin{IEEEbiography}[{\includegraphics[width=1in,height=1.25in,clip,keepaspectratio]{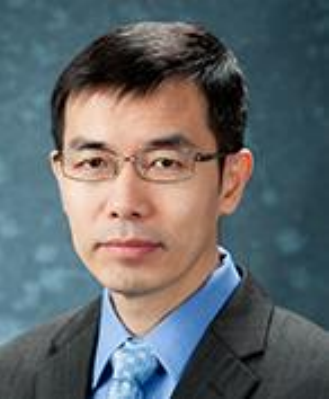}}]{Xiaoou Tang}
received the BS degree from the University of Science and Technology of China, Hefei, in 1990, and the MS degree from the University of Rochester, Rochester, NY, in 1991. He received the PhD degree from the Massachusetts Institute of Technology, Cambridge, in 1996. He is a Professor of the Department of Information Engineering, Chinese University of Hong Kong. He worked as the group manager of the Visual Computing Group at the Microsoft Research Asia from 2005 to 2008. His research interests include computer vision, pattern recognition, and video processing. He received the Best Paper Award at the IEEE Conference on Computer Vision and Pattern Recognition (CVPR) 2009 and Outstanding Student Paper Award at the AAAI 2015. He has served as a program chair of the IEEE International Conference on Computer Vision (ICCV) 2009 and associate editor of the IEEE Transactions on Pattern Analysis and Machine Intelligence. He is an Editor-in-Chief of the International Journal of Computer Vision. He is a fellow of the IEEE.
\end{IEEEbiography}

\end{document}